\documentclass[review]{elsarticle}
\usepackage[a4paper, total={5.5in, 8.5in}]{geometry}

\journal{Pattern Recognition}

\usepackage{lineno,hyperref}
\modulolinenumbers[5]
\usepackage[utf8]{inputenc}
\usepackage{graphicx, epstopdf}
\usepackage{amssymb}
\usepackage{bbm}
\usepackage{amsmath}
\usepackage{stmaryrd} % Iverson bracket
\usepackage{relsize}
\usepackage{mathtools}
\usepackage{multirow}
\usepackage{adjustbox}	% - Compressing tables
\usepackage{booktabs}
\usepackage{url}
\usepackage{hyperref}
\usepackage{pifont}% http://ctan.org/pkg/pifont

\usepackage{arydshln}

\usepackage{xcolor}
\definecolor{MRHC_color}{RGB}{221,0,221}
\definecolor{MChen_color}{RGB}{0,0,221}
\definecolor{MRSP1_color}{RGB}{221,0,0}
\definecolor{MRSP2_color}{RGB}{0,221,0}
\definecolor{MRSP3_color}{RGB}{85,85,85}
\definecolor{statV_color}{RGB}{0,150,0}
\definecolor{statX_color}{RGB}{150,0,0}
\newcommand{\BlackDiamond}{\tiny{\rotatebox[origin=c]{45}{$\blacksquare$}}}

\newcommand{\symMRSPOne}{\textcolor{MRSP1_color}{$\blacktriangle$}}
\newcommand{\symMRSPTwo}{\textcolor{MRSP2_color}{$\blacktriangle$}}
\newcommand{\symMRSPThree}{\textcolor{MRSP3_color}{$\blacktriangle$}}
\newcommand{\symMChen}{\textcolor{MChen_color}{$\BlackDiamond$}}

\newcommand{\statV}{\textcolor{statV_color}{\ding{51}}}
\newcommand{\statX}{\textcolor{statX_color}{\ding{55}}}

\usepackage[ruled,lined,linesnumbered,commentsnumbered,longend]{algorithm2e}
\usepackage{comment}
\usepackage{algpseudocode}
\usepackage{bm}

\usepackage{dutchcal}
\usepackage{caption}
\usepackage{subcaption}

\usepackage{xcolor,colortbl}

\begin{document}\sloppy
\begin{frontmatter}

\title{Multilabel Prototype Generation for Data Reduction in k-Nearest Neighbour classification}

\author[1]{Jose J. Valero-Mas\texorpdfstring{\corref{cor1}}{}}
\cortext[cor1]{Corresponding author: 
University Institute for Computer Research, University of Alicante, Carretera San Vicente del Raspeig s/n, 03690 Alicante, Spain. Tel.: +349-65-903772}
\ead{jjvalero@dlsi.ua.es}
\author[1]{Antonio Javier Gallego}
\author[2]{Pablo Alonso-Jim{\'e}nez}
\author[2]{Xavier Serra}

\address[1]{University Institute for Computer Research, University of Alicante, Spain}
\address[2]{Music Technology Group, Universitat Pompeu Fabra, Spain}

\begin{abstract}
Prototype Generation (PG) methods are typically considered for improving the efficiency of the $k$-Nearest Neighbour ($k$NN) classifier when tackling high-size corpora. Such approaches aim at generating a reduced version of the corpus without decreasing the classification performance when compared to the initial set. Despite their large application in multiclass scenarios, very few works have addressed the proposal of PG methods for the multilabel space. In this regard, this work presents the novel adaptation of four multiclass PG strategies to the multilabel case. These proposals are evaluated with three multilabel $k$NN-based classifiers, 12 corpora comprising a varied range of domains and corpus sizes, and different noise scenarios artificially induced in the data. The results obtained show that the proposed adaptations are capable of significantly improving---both in terms of efficiency and classification performance---the only reference multilabel PG work in the literature as well as the case in which no PG method is applied, also presenting statistically superior robustness in noisy scenarios. Moreover, these novel PG strategies allow prioritising either the efficiency or efficacy criteria through its configuration depending on the target scenario, hence covering a wide area in the solution space not previously filled by other works.
\end{abstract}


\end{frontmatter}

%\linenumbers

%-------------------------------------------------------------------------------------------------
\section{Introduction}
\label{sec:introduction}

% kNN
The $k$-Nearest Neighbour ($k$NN) classifier represents one of the most well-known algorithms for non-parametric supervised classification, mostly due to its conceptual simplicity and good statistical error properties~\citep{hart2000pattern}. For a given query, this method hypothesises about its category by querying the $k$ nearest neighbours of a reference corpus, following a specified similarity measure~\citep{bishop2006pattern}. In this regard, this classification strategy has been largely considered in a wide range of disparate fields as, for instance, diabetes detection~\cite{SUYANTO2022116857}, musical key estimation~\cite{george2022development} or handwritten signature verification~\cite{hancer2021wrapper}, among others.

% Limitations
However, as a representative case of the \textit{lazy} learning paradigm, $k$NN does not derive a model out of the reference corpus~\citep{mitchell1997machine}. In contrast, for every query, this method requires an exhaustive search among the elements of the aforementioned corpus, thus entailing low-efficiency figures in both classification time and memory usage~\citep{deng2016efficient}. Note that, while this inefficiency issue may be obviated in scenarios with limited amounts of data, when considering large data collections, $k$NN becomes intractable~\citep{GALLEGO2022108356}.

% Ways of improving kNN efficiency -> Highlight DR and PG
% In this context, 
Data Reduction (DR) stands as one of the most popular approaches in the related literature for tackling this drawback~\citep{garcia2015data}. This group of methods aims to reduce the size of the reference set for improving the efficiency of the model while keeping---or even increasing---the classification performance obtained with the original data. Among them, the Prototype Generation (PG) family represents one of the most competitive alternatives due to its remarkable reduction capabilities compared to other DR strategies~\cite{ESCALANTE2016569}. In a broad sense, PG derives an alternative reference set for the classifier by performing different selection and merging operations on the elements of the initial corpus following certain heuristics~\citep{triguero2012taxonomy}.

% Most of this work done for binary/multiclass classification; very scarce work done for multilabel
Due to the relevance of PG for efficient $k$NN-based classification, a considerable amount of research effort has been invested in proposing novel strategies as well as improving the existing ones~\citep{nanni2011prototype}. However, such research works have typically addressed \textit{multiclass} scenarios---classification tasks in which every single query is assigned to one category out of a set of mutually excluding labels---, hence neglecting the more general \textit{multilabel} scenario---case in which an undetermined number of categories is assigned to each query~\citep{zhang2013review}.

% M-RHC:
The work by Ougiaroglou et al.~\cite{Ougiaroglou:MRHC} represents one of the scarce works of a PG strategy devised to address multilabel scenarios. More precisely, this work proposes the adaptation of the state-of-the-art Reduction through Homogeneous Clustering (RHC) method~\citep{ougiaroglou2012efficient} to the multilabel space, obtaining the so-called Multilabel Reduction through Homogeneous Clustering (MRHC). The authors not only conclude that such adaptation remarkably improves the efficiency of the $k$NN classifier in multilabel scenarios but also state the need for contriving multilabel PG strategies due to the shortage of existing alternatives.

% Our proposal:
In this context, the present work further explores the proposal and use of PG methods for improving the efficiency of $k$NN classification in multilabel scenarios. More precisely, we introduce the novel adaptation of four PG strategies from their original multiclass formulation to the multilabel case. These proposals have been comprehensively evaluated considering several multilabel classification approaches based on $k$NN with a wide variety of corpora. Additionally, different percentages of label-level noise particularly devised for this multilabel framework---artificial alterations of the classes or labels of the data---have been induced in the corpora to assess the robustness of the proposals and their capability of dealing with such adverse scenarios. The results obtained report a statistically significant improvement in terms of both reduction capabilities and classification performance for all scenarios and noise levels contemplated compared to the exhaustive search carried out by the base $k$NN method and the reference MRHC reduction approach. In this regard, these novel proposals not only fill a gap in the scarce multilabel PG literature but also reportedly outperform the only existing strategy in the field, the commented MRHC algorithm.

% Roadmap:
The rest of the work is structured as follows: Section~\ref{sec:background} provides the theoretical background of the work; Section~\ref{sec:method} presents the proposed PG methods; Section~\ref{sec:experimental_setup} introduces the experimental setup; Section~\ref{sec:results} shows and discusses the results; and finally, Section~\ref{sec:conclusions} concludes the work and poses future research lines to pursue.

%-------------------------------------------------------------------------------------------------
\section{Background} 
\label{sec:background}

% Multiclass definition
To adequately describe multilabel classification, we initially introduce the multiclass framework, as it conceptually represents a simpler task. Formally, let $\mathcal{X}\in\mathbb{R}^{f}$ denote an $f$-dimensional feature space and $\mathcal{Y}_{mc}$ a set of discrete labels. Additionally, let $\mathcal{T}_{mc} = \left\{(\bm{x}_{i},y_{i}): \bm{x}_{i}\in\mathcal{X}, y_{i}\in\mathcal{Y}_{mc}\right\}_{i=1}^{|\mathcal{T}_{mc}|}$ represent an annotated collection of data where each datum $\bm{x}_{i}\in\mathcal{X}$ is related to label $y_{i}\in\mathcal{Y}_{mc}$ by an underlying function $h_{mc} : \mathcal{X}\rightarrow \mathcal{Y}_{mc}$. The goal of multiclass classification is retrieving the most accurate approximation $\hat{h}_{mc}\left(\cdot\right)$ to that underlying function.

% Multiclass kNN
Among the different alternatives for performing such an approximation task, the well-known $k$NN stands as one of the most common choices given its relevance in the Pattern Recognition field~\citep{GALLEGO2018531}. Formally, given a query $q\in\mathcal{X}$, this method models $\hat{h}_{mc}$ as:
\begin{equation}
    \hat{h}_{mc}\left(q\right) = \mbox{mode}\left(\mathcal{Y}_{mc}^{k}\left(\underset{\bm{x}_{i}\in\mathcal{T}_{mc}}{\arg\min}\prescript{}{k}{\left\{d\left(q,\bm{x}_{i}\right)\right\}}\right)\right)
\end{equation}

\noindent where $k$ stands for the number of neighbours considered, $d : \mathcal{X}\times\mathcal{X}\rightarrow\mathbb{R}^{+}_{0}$ is a dissimilarity measure, $\mbox{mode} : \mathcal{Y}_{mc}\rightarrow\mathcal{Y}_{mc}$ denotes the mode operator, and $\mathcal{Y}_{mc}^{k}$ is the set of labels retrieved from the closest $k$ elements to the query $q$.

% Change to multilabel (generalisation of the multiclass case)
As previously introduced, the multilabel paradigm constitutes a generalisation of the multiclass framework in which each individual instance may be associated with more than a single label~\citep{BelloNapolesVanhoofBello:IDA}. Formally, the set of multilabel data $\mathcal{T}_{ml} = \left\{(\bm{x}_{i},\mathbcal{y}_{i}): \bm{x}_{i}\in\mathcal{X}, \mathbcal{y}_{i}\subseteq\mathcal{Y}_{ml}\right\}_{i=1}^{|\mathcal{T}_{ml}|}$ relates datum $\bm{x}_{i}\in\mathcal{X}$ to a subset of classes $\mathbcal{y}_{i}\subseteq\mathcal{Y}_{ml}$, namely labelset, where $\mathcal{Y}_{ml} = \left\{\lambda_{1},\lambda_{2},\ldots,\lambda_{L}\right\}$ is an $L$-size collection of mutually non-exclusive labels~\citep{moyano2018review}. As in the multiclass case, the goal is retrieving the most accurate approximation $\hat{h}_{ml}\left(\cdot\right)$ to the underlying function $h_{ml} : \mathcal{X} \rightarrow \mathcal{Y}_{ml}$.

% Bridging MC and ML; focusing on kNN
To leverage the advantages of multiclass classifiers in multilabel scenarios, the literature considers two main approaches~\citep{GibajaVentura:ACM:Multilabel}: \textit{problem transformation} and \textit{algorithm adaptation}. We now describe these paradigms and report some commonly considered methods within them for $k$NN schemes as it represents the focus of the work. 

% Problem transformation
The \textit{problem transformation} paradigm disentangles the multilabel task into several single-label problems for then applying a multiclass $k$NN-based strategy for performing the classification task. Some of the most common alternatives are: the \textit{Binary Relevance} $k$NN (BR$k$NN), which decomposes the task into $L$ independent binary classification problems~\cite{zhang2018binary}; the \textit{Label Powerset} $k$NN (LP-$k$NN), which derives an alternative single-label corpus where each labelset is considered as a different class~\citep{RASTIN2021107526}; and Random $k$-Labelsets (RA$k$EL), which divides the initial set of labels into a number of small random subsets for then performing LP-$k$NN and creating an ensemble-based classifier~\cite{tsoumakas2010random}.

% Active k-labelsets ensemble for multi-label classification (ACkEL):  https://www.sciencedirect.com/science/article/pii/S0031320320303861?casa_token=GeIZmzrPo_EAAAAA:L6AIud0CeDnPNgyRIbDXWKBN4g_JgTIYWMHfAZ7Jvap3qWfxdgIjP0fpsnhVqBXpg9M7RteFpS8#bib0056

% Algorithm adaptation
In contrast, the \textit{algorithm adaptation} approach focuses on modifying the base multiclass classifier to fit the multilabel scenario. In this regard, the \textit{Multilabel} $k$NN (ML-$k$NN) proposed by Zhang and Zhou~\cite{zhang2007ml} expands the base $k$NN method resorting to a maximum-a-posteriori principle to determine the labelset of the query based on its neighbouring instances. Some extensions to this approach are the \textit{Dependent ML-$k$NN}~\cite{younes2008multi}, which models the different dependencies among the set of labels, the IBLR-ML method~\cite{cheng2009combining}, which expands the base ML-$k$NN one by combining it with logistic regression, or the combination of ensembles and ML-$k$NN as in the work by Zhu et al.~\cite{ZHU2021106933}.

% Efficiency issue in kNN
Nevertheless, while these transformations and adaptations allow the use of $k$NN in multilabel classification tasks, the inherent efficiency issue of these classifiers has been neglected in the literature. Note that, while some multilabel schemes such as the ML-$k$NN depict similar inefficiency figures to that of the multiclass $k$NN formulation since they explore the entire reference $\mathcal{T}_{ml}$ set, the BR$k$NN case is of particular relevance as it requires iterating through the $\mathcal{T}_{ml}$ set $L$ different times.

% PG for the win; not addressed in ML
The Prototype Generation (PG) family of methods stands as one of the most successful approaches for efficient $k$NN classification in multiclass cases~\citep{garcia2015data}. As a representative case of DR strategy, PG aims to obtain an alternative set $\mathcal{R}_{mc}$ by performing certain combinations and transformations on the elements of $\mathcal{T}_{mc}$ so that $|\mathcal{R}_{mc}| < |\mathcal{T}_{mc}|$ while keeping---or even improving---the classification performance. However, as aforementioned, to the best of our knowledge, there is a remarkable lack of methods for performing PG in multilabel scenarios. The sole exception to this assertion is the work by Ougiaroglou et al.~\cite{Ougiaroglou:MRHC} where the state-of-the-art multiclass PG method RHC was adapted to the multilabel space. In that work, the authors experimentally proved the usefulness of their PG proposal to improve the efficiency of the multilabel classification and stated the need for devising other alternatives to fill this existing gap in the literature.

% Proposal
In this context, the present work proposes a novel adaptation to the multilabel space of four well-known multiclass PG algorithms. More precisely, we consider the classic Chen reduction algorithm~\citep{chen1996sample} as well as the three different versions of the reference Reduction through Space Partitioning (RSP) strategy by S{\'a}nchez~\cite{sanchez2004high}. For this first-time adaptation to the multilabel space of such PG algorithms, this work proposes several mechanisms for both partitioning and integrating the labels of the multilabel prototypes of the initial corpus for eventually generating the instances of the reduced multilabel set. These novel methods are thoroughly compared, in terms of both performance and efficiency, to the state-of-the-art proposal by Ougiaroglou et al.~\cite{Ougiaroglou:MRHC} and to the case in which no reduction is performed considering different multilabel $k$NN-based classifiers, corpora, and noise scenarios. Such a study shall provide insights on whether the proposed multilabel PG methods cope with the commented efficiency issue without decreasing the classification performance and on their robustness as well as data cleansing capabilities in cases depicting the presence of noise in the data.

%-------------------------------------------------------------------------------------------------
\section{Prototype Generation in the multilabel space} 
\label{sec:method}

This section presents the proposed PG methods for the multilabel space. As commented, we focus on the first-time adaptation of four algorithms originally devised for multiclass cases: the Chen method~\citep{chen1996sample} and the three versions of the Reduction through Space Partitioning (RSP) strategy~\citep{sanchez2004high}. In this regard, the first part of the section introduces the original multiclass formulations of these algorithms and the second one presents their respective multilabel adaptations proposed in this work.

\subsection{Reference multiclass PG}

The considered multiclass PG strategies---the Chen method as well as the different RSP versions---constitute representative examples of the so-called \textit{space splitting} policy~\citep{CastellanosValeroMasCalvoZaragoza:SOCO:PGString}, which typically follows a two-step approach: a first stage, \textit{space partitioning}, divides the feature space of the multiclass set $\mathcal{T}_{mc}$ into different regions using certain heuristics; after that, the \textit{prototype merging} stage computes new prototypes from each region attending to different criteria, producing the reduced set $\mathcal{R}_{mc}$. The existing PG strategies under this framework, therefore, essentially differ in the particular splitting and prototype generation heuristics considered.

\begin{figure}[!ht]
	\centering
	\begin{subfigure}[t]{0.45\textwidth}
		\centering
		\includegraphics[width=.9\textwidth]{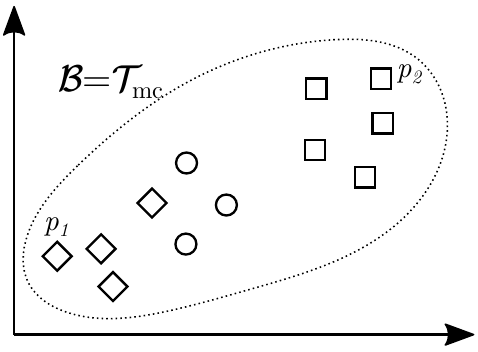}
		\caption{Space partitioning stage when $n_c=1$.}
		\label{fig:chen1}
	\end{subfigure}
	\qquad
	\begin{subfigure}[t]{0.45\textwidth}
		\centering
		\includegraphics[width=.9\textwidth]{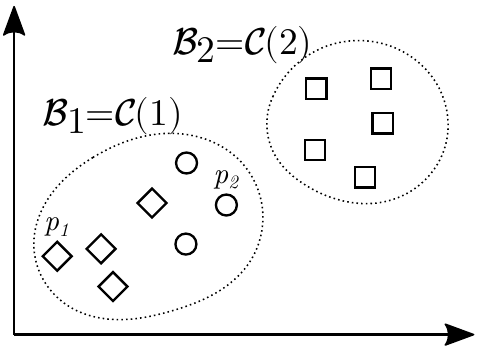}
		\caption{Space partitioning stage when $n_c=2$.}
		\label{fig:chen2}
	\end{subfigure}
	\begin{subfigure}[t]{0.45\textwidth}
		\centering
		\includegraphics[width=.9\textwidth]{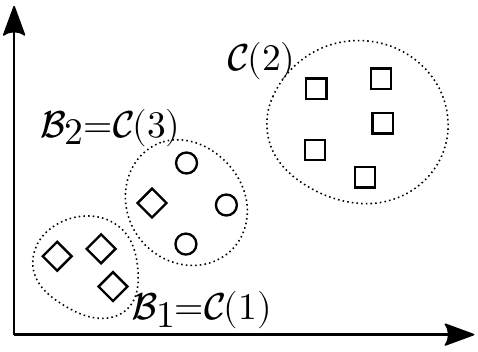}
		\caption{Space partitioning stage when $n_c=3$.}
		\label{fig:chen3}
	\end{subfigure}
	\qquad
	\begin{subfigure}[t]{0.45\textwidth}
		\centering
		\includegraphics[width=.9\textwidth]{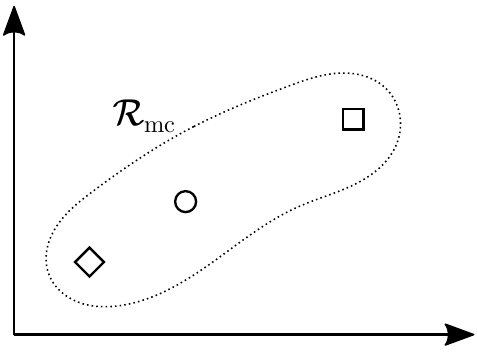}
		\caption{Prototype merging phase.}
		\label{fig:chen4}
	\end{subfigure}
	\caption{Graphical illustration of the multiclass Chen PG method. The example depicts the results of the space partitioning process (cases~\ref{fig:chen1} to \ref{fig:chen3}) and the prototype merging phase (case~\ref{fig:chen4}) when considering $n_d=3$ subsets. Symbols $p_{1}$ and $p_{2}$ denote the two furthest prototypes in the cluster to be divided.}
\label{fig:Multiclass_Chen}
\end{figure}

In the specific case of the Chen and RSP PG families, the aforementioned heuristics depict some similarities. In this regard, we first present the particular approach followed by the Chen method in Algorithm~\ref{alg:ChenMethod}, aided by the graphical illustration in Figure~\ref{fig:Multiclass_Chen}, for then commenting on the different points on which the three RSP strategies differ from it.

\begin{algorithm}[!ht]
\caption{Chen algorithm for multiclass PG~\citep{chen1996sample}}
\label{alg:ChenMethod}
\linespread{1.25}\selectfont
\SetAlgoLined
\SetKwInOut{KwIn}{Input}
\SetKwInOut{KwOut}{Output}
\KwIn{
$\mathcal{T}_{mc}\subset\mathcal{X}\times\mathcal{Y}_{mc} \leftarrow$ Multiclass corpus\\
$ n_{d} \leftarrow$ Number of resulting partitions \\
$d(\cdot,\cdot) \leftarrow$ Dissimilarity measure\\
}
\KwOut{$\mathcal{R}_{mc} \leftarrow$ Reduced set} 
\textbf{Let} $n_{c}=i=1$, $\mathcal{C}_{mc} = \emptyset$, $\mathcal{B} = \mathcal{T}_{mc}$  \Comment{\textbf{Space partitioning}}\\
\textbf{Let} $p_{1}, p_{2}$ be the farthest prototypes in $\mathcal{B}$\\
\While  {$n_{c} < n_{d}$}  { 
\textbf{Divide} $\mathcal{B}$ into subsets:\\
\hspace{1em}$\mathcal{B}_{1} = \left\{ p \in \mathcal{B} : d(p,p_{1}) \leq d(p, p_{2})\right\}$\\
\hspace{1em}$\mathcal{B}_{2} = \left\{ p \in \mathcal{B} : d(p,p_{1}) > d(p, p_{2})\right\}$\\
\textbf{Set} $n_{c} = n_{c} + 1$, $\mathcal{C}_{mc}(i) = \mathcal{B}_{1}$, and $\mathcal{C}(n_{c}) = \mathcal{B}_{2}$\\
\textbf{Divide} $\mathcal{C}_{mc}$ into subsets:\\
\hspace{1em}$\mathcal{I}_{1} = \left\{i : \left|\left\{y\in C_{mc}(i)\right\}\right| > 1\right\}$\\
\hspace{1em}$\mathcal{I}_{2} = \left\{j : j \leq n_{c}\right\} - \mathcal{I}_{1}$\\
\textbf{Let} $\mathcal{I} = \mathcal{I}_{1}$ if $\mathcal{I}_{1} \neq \emptyset$ else $\mathcal{I}_{2}$\\
\textbf{Find} farthest points $q_{1}(i), q_{2}(i)$ in $\mathcal{C}_{mc}(i)~\forall~i \in \mathcal{I}$\\
\textbf{Let} $j = \arg\max_{j\in\left[1,i\right]} d\left(q_{1}\left(j\right), q_{2}\left(j\right)\right)$\\
\textbf{Set} $\mathcal{B} = \mathcal{C}_{mc}(j)$, $p_{1} = q_{1}(j)$, and $p_{2} = q_{2}(j)$
}
\textbf{Compute} $\mathcal{R}_{mc} = \left\{(\bm{x}_{i},y_{i})\right\}_{i=1}^{n_d}$ as:\Comment{\textbf{Prototype merging}}\\
\hspace{1em}$\bm{x}_{i} = \mbox{median}\left(\left\{\bm{x}\in\mathcal{C}_{mc}\left(i\right)\right\}\right)$\\
\hspace{1em}$y_{i} = \mbox{mode}\left(\left\{y\in\mathcal{C}_{mc}\left(i\right)\right\}\right)$
\end{algorithm}

As it may be observed in the algorithm, the method iteratively divides the feature space of $\mathcal{T}_{mc}$ into $n_{d}$---user parameter---disjoint subsets which are denoted as $\mathcal{C}_{mc}(i)$ where $\bigcup_{i=1}^{n_d}\mathcal{C}_{mc}(i) = \mathcal{T}_{mc}$. For that, the largest subset in each iteration is divided in two attending to the distance between the two farthest prototypes in it. Eventually, for each of the $n_{d}$ regions, a new prototype is obtained as the median of the features of the elements in it and labelled after the most common class. Hence, the size of the resulting reduced set equals the number of partitions selected by the user, \textit{i.e.}, $|\mathcal{R}_{mc}|=n_{d}$. 

The RSP family, as commented, builds upon Chen's proposal by modifying some of the space partitioning and/or prototype merging stages. The first RSP version---RSP1---considers the Chen algorithm prone to discard underrepresented classes due to its prototype merging policy (lines 16-18 in Algorithm~\ref{alg:ChenMethod}). Thus, instead of computing a single prototype for each of the $n_{d}$ regions and labelling them after the most represented class in each partition, RSP1 only merges prototypes sharing the same label. Hence, each region is now represented by as many prototypes as the number of classes it contains. In this case, therefore, the size of the reduced set may not be known in advance but accomplishes $|\mathcal{R}_{mc}|\geq n_{d}$.

The second version of RSP---RSP2---expands RSP1 by modifying the criterion for selecting the region to split (lines 12-13 in Algorithm~\ref{alg:ChenMethod}). RSP2 considers the overlapping degree criterion, which is defined as the ratio of the average distance between instances belonging to different classes and the average distance between instances that are from the same class. The region with the largest overlapping degree is the one to be divided.

The third RSP reduction heuristic---RSP3---is based on the idea that each resulting region should represent a cluster of instances belonging to only one class. Thus, this approach modifies the Chen method so that it iteratively performs the space partitioning stage (line 3 in Algorithm~\ref{alg:ChenMethod}) until all resulting sets are homogeneous in terms of class representation, remaining the prototype merging phase of the algorithm unaltered. Hence, unlike the RSP1 and RSP2 strategies, the RSP3 approach does not require the $n_d$ parameter related to the number of resulting regions since the method exclusively relies on this class homogeneity criterion to accomplish the space partitioning stage.

\subsection{Multilabel PG proposals}

Having introduced the four reference PG methods in their multiclass formulation, we now present their respective proposed multilabel adaptations.

The multilabel space splitting PG framework may be formulated in an analogous manner to that of the multiclass case. Initially, the \textit{space partitioning} phase divides the multilabel set $\mathcal{T}_{ml}\subset\mathcal{X}\times\mathcal{Y}_{ml}$ into $n_{d}$ non-overlapping multilabel regions $C_{ml}$ such that $\bigcup_{i=1}^{n_d}\mathcal{C}_{ml}(i)=\mathcal{T}_{ml}$. After the convergence of this stage, the \textit{prototype merging} step retrieves the multilabel set of data $\mathcal{R}_{ml}$ generated out of these $C_{ml}$ clusters by following a certain approach, where $|\mathcal{R}_{ml}| \leq |\mathcal{T}_{ml}|$. Within this framework, we introduce the different modifications proposed for accommodating the presented multiclass PG methods to such a scenario.

Our first proposal is the adaptation of the Chen algorithm, namely \textit{Multilabel Chen} or \textit{MChen}. Since the space partitioning stage (lines 1-15) computes the set of clusters $\mathcal{C}_{mc}$ only relying on the set of features $\mathcal{X}$, no adaptation is required for its multilabel formulation to obtain set $\mathcal{C}_{ml}$. Oppositely, given that the prototype merging stage (lines 16-18) usually requires combining elements from different classes, the question arises about the proper approach to do so in multilabel spaces since the simple selection of the most common label in the $\mathcal{C}_{ml}$ cluster is not suitable for the considered scenario. 

\begin{figure}[!ht]
	\centering
	\begin{subfigure}[t]{0.4\textwidth}
		\centering
		\includegraphics[width=.9\textwidth]{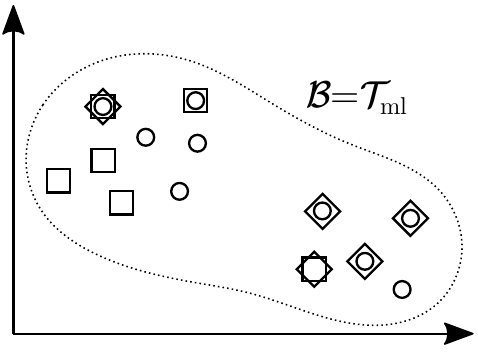}
		\caption{Example of an initial set of multilabel data $\mathcal{T}_{ml}$.}
		\label{fig:ml_initial_space}
	\end{subfigure}
	\qquad
	\begin{subfigure}[t]{0.4\textwidth}
		\centering
		\includegraphics[width=.9\textwidth]{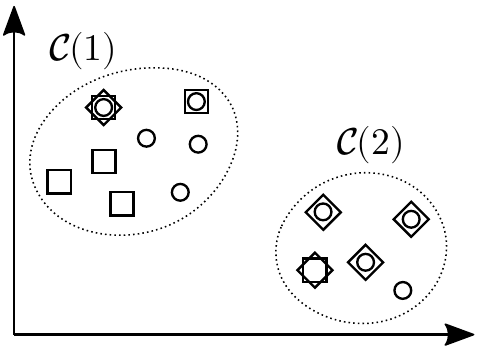}
		\caption{Space partitioning result considered for the rest of examples on the elements in Figure~\ref{fig:ml_initial_space} with $n_d=2$.}
		\label{fig:ml_space_part_result}
	\end{subfigure}
	\\
	\begin{subfigure}[t]{0.4\textwidth}
		\centering
		\includegraphics[width=.9\textwidth]{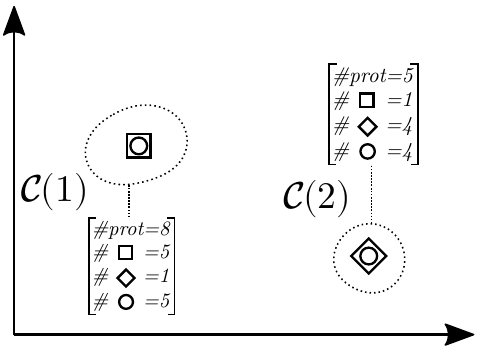}
		\caption{Outcome of the prototype merging procedure ($\mathcal{R}_{ml}$) by the MChen proposal on the result in Figure~\ref{fig:ml_space_part_result}.}
		\label{fig:mchen}
	\end{subfigure}
	\qquad
	\begin{subfigure}[t]{0.4\textwidth}
		\centering
		\includegraphics[width=.9\textwidth]{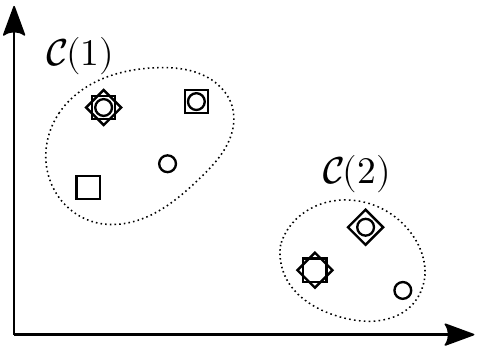}
		\caption{Outcome of the prototype merging procedure ($\mathcal{R}_{ml}$) by the proposed MRSP1 on the result in Figure~\ref{fig:ml_space_part_result}.}
		\label{fig:mrsp1}
	\end{subfigure}
	\begin{subfigure}[t]{0.4\textwidth}
		\centering
		\includegraphics[width=.9\textwidth]{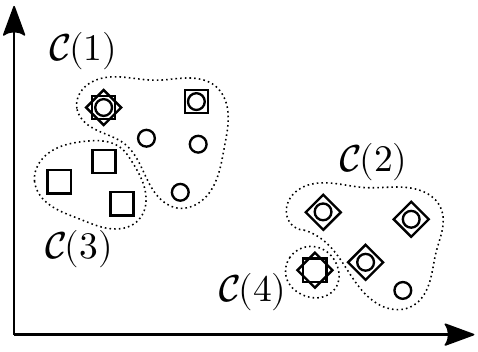}
		\caption{Result of the space partitioning phase obtained by the MRSP3 proposal on the prototypes in Figure~\ref{fig:ml_initial_space}.}
		\label{fig:mrsp3_part}
	\end{subfigure}
	\qquad
	\begin{subfigure}[t]{0.4\textwidth}
		\centering
		\includegraphics[width=.9\textwidth]{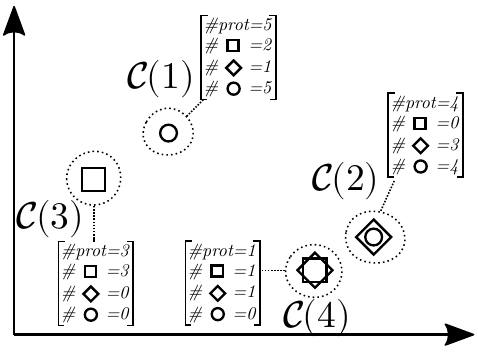}
		\caption{Outcome of the prototype merging procedure ($\mathcal{R}_{ml}$) by the proposed MRSP3 on the result in Figure~\ref{fig:mrsp3_part}.}
		\label{fig:mrsp3_merging}
	\end{subfigure}
	\caption{Graphical illustration of the multilabel PG proposals introduced in the work. Figure~\ref{fig:ml_initial_space} represents a multilabel set of train data $\mathcal{T}_{ml}$ to be reduced. Figure~\ref{fig:ml_space_part_result} shows the space partitioning results on which the different reduction proposals are based, except for the MRSP3 one, whose case is illustrated in Figure~\ref{fig:mrsp3_part}. Prototype merging graphs \ref{fig:mchen} and \ref{fig:mrsp3_merging} depict the number of prototypes---denoted as \textit{\#prot}---and the cardinality of labels---$\#\square$, $\#\circ$, and $\#\diamond$---for each of the original clusters.}
\label{fig:chen}
\end{figure}

In this regard, we resort to the policy devised by Ougiaroglou et al.~\cite{Ougiaroglou:MRHC} for the MRHC method in which the resulting prototype keeps the labels present in at least half of the instances of the cluster. Mathematically, the labelset assigned to the resulting element in cluster $\mathcal{C}_{ml}(i)$ is given by:
\begin{equation}
\mathbcal{y}_{i} = \left\{\lambda : |\mathcal{C}_{ml}(i)|_{\lambda}\geq\frac{|\mathcal{C}_{ml}(i)|}{2} \;\;\forall~\lambda\in\mathcal{C}_{ml}(i)\right\}
\label{eq:ProtGenMChen}
\end{equation}
%  = |\{\lambda\} \cap \mathcal{C}_{ml}(i)|

\noindent where $|\mathcal{C}_{ml}(i)|_{\lambda}$ denotes the cardinality of label $\lambda$ in subset $\mathcal{C}_{ml}(i)$. This expression replaces that in line 18 of Algorithm~\ref{alg:ChenMethod} whereas the policy followed for obtaining the set of features (line 17) is not modified. Figure~\ref{fig:mchen} provides a graphical example of this merging procedure considering the space partitioning result shown in Figure~\ref{fig:ml_space_part_result}.

The second proposal is the \textit{Multilabel RSP1} or \textit{MRSP1}. As aforementioned, the RSP1 states that, during the prototype merging stage and for each cluster $\mathcal{C}_{mc}(i)$, one prototype must be retrieved for each class present in it. The MRSP1 adapts such stage by resorting to a labelset approach (lines 16-18), \textit{i.e.} each labelset is considered a different class and the instances with the same labelset are merged and assigned to it. Mathematically, set $\mathcal{R}_{ml}$ is obtained as:
\begin{equation}
    \mathcal{R}_{ml} = \left\{\left(\mbox{median}\left(\left\{\bm{x}_{j}:(\bm{x}_{j},\mathbcal{y}_{j})\in\mathcal{C}_{ml}(i), \mathbcal{y}_{j} = \mathbcal{y}_{k}\right\}\right), \mathbcal{y}_{k}\right)\right\}_{i=1}^{n_{d}}
    \label{eq:ProtGenMRSP1}
\end{equation}

\noindent where $k = \left|\left\{\mathbcal{y}\in\mathcal{C}_{ml}(i)\right\}\right|$ is the number of labelsets in the $i$-th cluster $\mathcal{C}_{ml}(i)$ and $j\in\left[1,|\mathcal{C}_{ml}(i)|\right]$. Figure~\ref{fig:mrsp1} provides a graphical example of this procedure based on the space partitioning result depicted in Figure~\ref{fig:ml_space_part_result}.

The \textit{Multilabel RSP2} or \textit{MRSP2} proposal generalises the space partitioning approach based on the overlapping degree from the RSP2 method to the multilabel space (lines 12-13). For that, as in the MRSP1 proposal, we resort to a labelset approach: each labelset is considered a different class and the overlapping degree $\Phi_{i}$ of the $i$-th $C_{ml}(i)$ region is computed as the ratio of the average distance between instances belonging to different labelsets---$D^{\neq}$---and the average distance between instances of the same labelset---$D^{=}$.

In formal terms, for the $i$-th region, these pairwise distance values $D^{\neq}$ and $D^{=}$ are respectively computed as:
\begin{eqnarray}
D^{\neq} &=& \left\{d(\bm{x}_{j},\bm{x}_{k}) : (\bm{x}_{j},\bm{y}_{j})\wedge(\bm{x}_{k},\bm{y}_{k})\in\mathcal{C}_{ml}(i), j\neq k, \bm{y}_{j}\neq \bm{y}_{k}\right\}\\
D^{=} &=& \left\{d(\bm{x}_{j},\bm{x}_{k}) : (\bm{x}_{j},\bm{y}_{j})\wedge(\bm{x}_{k},\bm{y}_{k})\in\mathcal{C}_{ml}(i), j\neq k, \bm{y}_{j}=\bm{y}_{k}\right\}
\end{eqnarray}

\noindent with $1\leq j,\,k\leq n_{d}$. Based on this, the overlapping degree $\Phi_{i}$ for the same $i$-th region is eventually obtained as:
\begin{equation}
\Phi_{i} = \frac{\sum_{j=1}^{|D^{\neq}|}{D^{\neq}(j)}}{\sum_{k=1}^{|D^{=}|}{D^{=}(k)}}\cdot\frac{|D^{=}|}{|D^{\neq}|}
\end{equation}

Note that, after the convergence of the space partitioning stage, the prototype merging policy in Equation~\ref{eq:ProtGenMRSP1} introduced for MRSP1 is applied.

The last proposal is the \textit{Multilabel RSP3} or \textit{MRSP3}. In this case, we must generalise the cluster homogeneity concept of the RSP3 method to automatically estimate the $n_{d}$ number of clusters. For that, we resort to the criterion posed by Ougiaroglou et al.~\cite{Ougiaroglou:MRHC} which states that a set of multilabel data is considered to be homogeneous if there is, at least, one common label among all the prototypes in the set, \textit{i.e.} $\exists~\lambda\in\mathcal{C}_{ml}(i)\mbox{\hspace{.5em}s.t.}~ |\mathcal{C}_{ml}(i)|_{\lambda} = |\mathcal{C}_{ml}(i)|$. This substitutes the condition in line 3 in Algorithm~\ref{alg:ChenMethod} so that the process finishes when this homogeneity criterion is accomplished by all regions. After this space partitioning stage, the set of clusters $\mathcal{C}_{ml}$ is further processed following the prototype merging approach of the MChen proposal in Equation~\ref{eq:ProtGenMChen}.

Finally, Figures~\ref{fig:mrsp3_part} and \ref{fig:mrsp3_merging} respectively show the result of the space partitioning and prototype merging phases of the introduced MRSP3 proposal.

%-------------------------------------------------------------------------------------------------
\section{Experimental set-up} 
\label{sec:experimental_setup}

\begin{figure}[!ht]
    \centering
    \includegraphics[width=.9\textwidth]{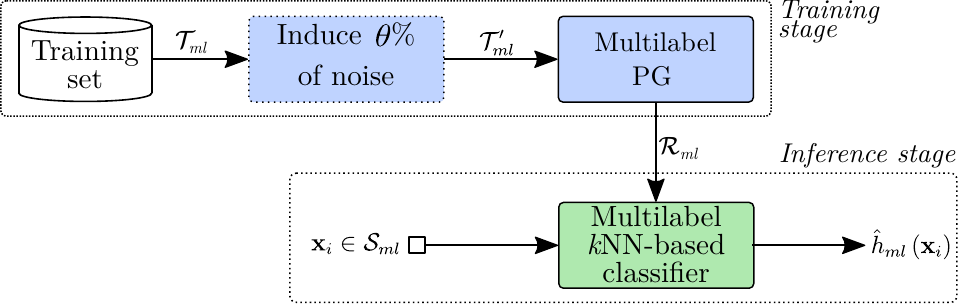}
    \caption{Experimental scheme for the comparative assessment of the PG methods.}
    \label{fig:experimental_scheme}
\end{figure}

This section presents the experimental scheme designed for comparatively assessing the proposed multilabel PG methods. For an easier description, this procedure is graphically illustrated in Figure~\ref{fig:experimental_scheme}.

During the training phase of the procedure, the set of train data $\mathcal{T}_{ml} \subset \mathcal{X}\times\mathcal{Y}_{ml}$ is altered to induce certain noise level in the instances controlled by the user parameter $\theta\in\left[0,1\right]$, retrieving set $\mathcal{T}_{ml}^{\prime}$. Then, this latter data collection $\mathcal{T}_{ml}^{\prime}$ is processed by a multilabel PG method to obtain a reduced version of the set---namely $\mathcal{R}_{ml}$---that is used as the reference set for the multilabel $k$NN-based classifier. It must be noted that the noise induction process represents an optional stage in the posed pipeline. Hence, as it will be shown, the first experimental part does not induce any noise by setting $\theta=0$ while the second one will analyse the robustness and data cleansing capabilities of the reduction methods to the data corruption process by considering $\theta > 0$.

During the inference stage, a test set of multilabel data $\mathcal{S}_{ml} \subset \mathcal{X}\times\mathcal{Y}_{ml}$ drawn from the same distribution as the train data $\mathcal{T}_{ml}$ but disjoint from it is considered for evaluating the method. Using a $\hat{h}_{ml}\left(\cdot\right)$ prediction function from the particular multilabel $k$NN-based classification strategy at hand, each sample $\bm{x}_{i}\in\mathcal{S}_{ml}$ is given a labelset that is eventually compared to that in the ground-truth based on certain evaluation criteria.

The remainder of the section presents the corpora used for assessing the multilabel PG proposals, the noise induction procedure used, the considered $k$NN-based classification strategies, and the contemplated evaluation protocol.

\subsection{Corpora}

We have considered 12 multilabel corpora from the Mulan repository~\citep{mulan} comprising a varied range of domains, corpus sizes, initial space dimensionalities, and target label spaces. The precise details in terms of size, features, and label dimensionality of these sets are provided in Table~\ref{tab:corpora}. Note that the \textit{cardinality}---average number of labels associated with each instance---and \textit{density}---ratio of cardinality and label dimensionality of the corpus---measures are provided for each corpus as they represent common descriptors in the multilabel classification field. In addition, we also provide the \textit{mean imbalance ratio} (\textit{MeanIR}) index that estimates the imbalance level of multilabel corpora and is obtained as:
\begin{equation}
    \mbox{\textit{MeanIR}}= \frac{1}{\left|\mathcal{Y}_{ml}\right|}\sum_{\lambda\in\left|\mathcal{Y}_{ml}\right|}\frac{\underset{\forall\lambda^{\prime}\in\mathcal{Y}_{ml}}{\max}\left(\mathlarger{\sum}_{i=1}^{\left|\mathcal{T}_{ml}\right|}\llbracket \lambda^{\prime}\in\mathbcal{y}_{i}\rrbracket\right)}{\mathlarger{\sum}_{i=1}^{\left|\mathcal{T}_{ml}\right|}\llbracket \lambda\in\mathbcal{y}_{i}\rrbracket}
\end{equation}
\noindent where the descriptor $\mbox{\textit{MeanIR}}\in\left[1,\infty\right)$ reports sharper imbalance rates as the value increases, and $\llbracket\cdot\rrbracket\rightarrow\left\{0,1\right\}$ represents the Iverson bracket, which outputs the unit value when the condition in the argument is met and zero otherwise.

\begin{table}[!ht]
    \caption{Summary of the corpora considered for the experimentation. Each corpus is described in terms of its data domain, partition sizes, dimensionality of input data (features) and output space (labels), cardinality, and density.}
    \label{tab:corpora}
    \centering
    \setlength{\tabcolsep}{4.5pt}
    \renewcommand{\arraystretch}{.85}
    \resizebox{\textwidth}{!}{%
    \begin{tabular}{lcccccccc}
\toprule[1pt]
\multirow{2}{*}{\textbf{Name}} & \multirow{2}{*}{\textbf{Domain}} &  \multicolumn{2}{c}{\textbf{Corpus size}} & \multicolumn{2}{c}{\textbf{Dimensionality}} & \multirow{2}{*}{\textbf{Cardinality}} & \multirow{2}{*}{\textbf{Density}} & \multirow{2}{*}{\textbf{MeanIR}}\\
\cmidrule(rl){3-4} \cmidrule(rl){5-6}
 & &    \textbf{Train} & \textbf{Test} &   \textbf{Features} ($f$) & \textbf{Labels ($L$)}\\
\cmidrule(rl){1-1} \cmidrule(rl){2-2} \cmidrule(rl){3-9}
Bibtex & Text &	\multicolumn{1}{r}{4,880} &	\multicolumn{1}{r}{2,515} &	\multicolumn{1}{r}{1,836} &	\multicolumn{1}{r}{159} &	2.40 &	0.015 & \phantom{0}12.78\\
Birds & Audio &	\multicolumn{1}{r}{322} &	\multicolumn{1}{r}{323} &	\multicolumn{1}{r}{260} &	\multicolumn{1}{r}{19} &	1.01 &	0.053 & \phantom{00}6.10\\
Corel5k & Image &	\multicolumn{1}{r}{4,500} &	\multicolumn{1}{r}{500} &	\multicolumn{1}{r}{499} &	\multicolumn{1}{r}{374} &	3.52 &	0.009 & 183.29\\
Emotions & Music &	\multicolumn{1}{r}{391} &	\multicolumn{1}{r}{202} &	\multicolumn{1}{r}{72} &	\multicolumn{1}{r}{6} &	1.87 &	0.311 & \phantom{00}1.49\\
Genbase	& Biology & \multicolumn{1}{r}{463} &	\multicolumn{1}{r}{199} &	\multicolumn{1}{r}{1,186} &	\multicolumn{1}{r}{27} &	1.25 &	0.046 & \phantom{0}31.60\\
Medical & Text &	\multicolumn{1}{r}{333} &	\multicolumn{1}{r}{645} &	\multicolumn{1}{r}{1,449} &	\multicolumn{1}{r}{45} &	1.25 &	0.028 & \phantom{0}48.59\\
rcvsubset1 & Text &	\multicolumn{1}{r}{3,000} &	\multicolumn{1}{r}{3,000} &	\multicolumn{1}{r}{47,236} &	\multicolumn{1}{r}{101} &	2.88 &	0.029 & 191.42\\
rcvsubset2 & Text &	\multicolumn{1}{r}{3,000} &	\multicolumn{1}{r}{3,000} &	\multicolumn{1}{r}{47,236} &	\multicolumn{1}{r}{101} &	2.63 &	0.026 & 177.89\\
rcvsubset3 & Text &	\multicolumn{1}{r}{3,000} &	\multicolumn{1}{r}{3,000} &	\multicolumn{1}{r}{47,236} &	\multicolumn{1}{r}{101} &	2.61 &	0.026 & 192.48\\
rcvsubset4 & Text &	\multicolumn{1}{r}{3,000} &	\multicolumn{1}{r}{3,000} &	\multicolumn{1}{r}{47,229} &	\multicolumn{1}{r}{101} &	2.49 &	0.025 & 170.84\\
Scene & Image &	\multicolumn{1}{r}{1,211} &	\multicolumn{1}{r}{1,196} &	\multicolumn{1}{r}{294} &	\multicolumn{1}{r}{6} &	1.07 &	0.179 & \phantom{00}1.33\\
Yeast & Biology &	\multicolumn{1}{r}{1,500} &	\multicolumn{1}{r}{917} &	\multicolumn{1}{r}{103} &	\multicolumn{1}{r}{14} &	4.24 &	0.303 & \phantom{00}7.27\\
\bottomrule[1pt]
    \end{tabular}
}
\end{table}

Note that, for the sake of reproducible research, we have used the partitions defined by Szyma{\'n}ski and Kajdanowicz in these particular corpora~\cite{ScikitML}.

\subsection{Noise induction procedure}\label{subsect:noise_induction}

To examine the actual robustness of both the existing and the proposed multilabel PG methods, we artificially introduce noise in the data. Note that, to our best knowledge, no previous work has assessed the robustness of multilabel PG methods by performing a noise induction process. Hence, we resort to the procedure by Natarajan et al.~\cite{natarajan2013learning} that is commonly considered in the multiclass PG literature: noise is introduced in the data by swapping the labels of pairs of prototypes randomly chosen from the train partition. Algorithm~\ref{alg:NoiseInduction} provides a formal description of the adaptation of this procedure to the multilabel space, in which the user parameter $\theta\in\left[0,1\right]$ represents the induced noise rate, \textit{i.e.}, the percentage of prototypes that change their label.

\begin{algorithm}[!ht]
\caption{Noise induction procedure}
\label{alg:NoiseInduction}
\linespread{1.25}\selectfont
\SetAlgoLined
\SetKwInOut{KwIn}{Input}
\SetKwInOut{KwOut}{Output}
\KwIn{
$\mathcal{T}_{ml}\subset\mathcal{X}\times\mathcal{Y}_{ml} \leftarrow$ Multilabel train corpus\\
$ \theta \leftarrow$ Noise level parameter \\
}
\KwOut{$\mathcal{T}_{ml}^{\prime}\subset\mathcal{X}\times\mathcal{Y}_{ml} \leftarrow$ Noisy multilabel train corpus\\} 
\textbf{Let} $\Theta=\left\{\left(\bm{x}_{i},\bm{y}_{i}\right)\right\}_{i=1}^{\theta\cdot|\mathcal{T}_{ml}|}\in_{R}\mathcal{T}_{ml}$  \Comment{Random sampling of set $\mathcal{T}_{ml}$}\\
\textbf{Let} $\mathcal{T}^{\prime}_{ml} = \mathcal{T}_{ml} - \Theta$\\
\For  {$i\in\left[0,\ldots, |\Theta|/2\right]$}  {
\textbf{Save} labelset of the $i$-th element in set $\Theta$: $\bm{y}^{\prime} = \bm{y}\in\Theta_{i}$\\
\textbf{Put} labelset in $|\Theta|-i$ in the $i$-th sample: $\bm{y}\in\Theta_{i} = \bm{y}\in\Theta_{|\Theta|-i}$\\
\textbf{Set} $\bm{y}^{\prime}$ as the labelset of the $|\Theta|-i$-th element: $\bm{y}\in\Theta_{|\Theta|-i} = \bm{y}^{\prime}$\\
}
\textbf{Let} $\mathcal{T}^{\prime}_{ml} = \mathcal{T}^{\prime}_{ml} \cup \Theta$
\end{algorithm}

As aforementioned, the particular case of $\theta=0$ represents that in which no noise is induced in the corpus, hence being $\mathcal{T}_{ml}^{\prime} = \mathcal{T}_{ml}$. In the subsequent experimentation, we will assess the proposals presented in this work considering both a noise-free scenario ($\theta=0$) as well as under different levels of induced noise typically considered in the related literature.

Note that, while this particular noise induction policy may be deemed simplistic, it constitutes a first approximation to assess the robustness of multilabel PG methods in the context of label-level distortions. Nevertheless, other procedures that contemplate the multilabel nature of these data may also provide some additional insights about the performance of these methods, such as swapping only part of the labels between pairs of instances, randomly including or eliminating classes for each prototype, or simply duplicating labelsets among elements in the corpus, and will be explored in future research.

\subsection{Classification strategies}

We have selected three reference multilabel techniques based on $k$NN as classification methods: BR$k$NN and LP-$k$NN from the transformation paradigm as well as ML-$k$NN based on the algorithm adaptation premise. In all cases, the Euclidean distance has been used as the dissimilarity measure.

Regarding the $k$ parameter representing the number of neighbours, we have considered the values $k\in\left\{1,3,5,7\right\}$. Note that this parameter is not optimised by any means during the experimentation since the aim is to examine its influence on the overall classification performance in relation to the PG mechanisms.

\subsection{Evaluation metrics}

To assess the goodness of the proposals, we consider two criteria: classification performance and efficiency figures. 

With respect to the former criterion, we resort to the Hamming Loss (HL) as it constitutes a commonly considered approach for measuring the goodness of multilabel classifiers~\citep{madjarov2012extensive}. This metric, which is defined as the fraction of the wrong predicted labels with respect to the total number of labels, can be mathematically posed as:
\begin{equation}
    \mbox{HL} = \frac{1}{\left|\mathcal{S}_{ml}\right|}\sum_{i=1}^{\left|\mathcal{S}_{ml}\right|}{\frac{1}{L}\cdot\left|\mathbcal{y}_{i}\;\Delta\;\hat{h}_{ml}\left(\bm{x}_{i}\right)\right|}
\end{equation}

\noindent where $\mathcal{S}_{ml} \subset \mathcal{X}\times\mathcal{Y}_{ml}$ denotes the multilabel set of test data, $\Delta$ is the symmetric difference of ground-truth $\mathbcal{y}_{i}$ and predicted $\hat{h}_{ml}\left(\bm{x}_{i}\right)$ sets, and $L$ is the number of labels.

As commonly done in the DR field, efficiency is assessed by comparing the size of the reduced set $\mathcal{R}_{ml}$ normalised by that of the training set $\mathcal{T}_{ml}$~\citep{RICOJUAN2019105803}. Computation time, which is typically discarded as an evaluation metric due to its variability depending on the load of the computing system, is additionally reported as a supplementary figure of merit for the analysis of the particular implementations provided in this work.

It must be noted that PG methods for $k$NN seek to simultaneously optimise two contradictory goals, set size reduction and classification performance, being not possible to achieve a global optimum. Hence, as in reference works from the literature~\citep{CALVOZARAGOZA20151608,valero2017experimental}, we address it as a Multi-objective Optimisation Problem in which the two aforementioned objectives are meant to be optimised. The different solutions under this framework---there may exist more than one---are retrieved by resorting to the concept of non-dominance: one solution is said to dominate another if it is better or equal in each goal function and, at least, strictly better in one of them. Those elements, typically known as non-dominated, constitute the Pareto frontier in which all elements are deemed as optimal solutions without any order among them.

%-------------------------------------------------------------------------------------------------
\section{Results}\label{sec:results}

This section introduces and discusses the results obtained by the proposed multilabel PG methods with the evaluation methodology considered. For comparison purposes, the reference MRHC method and the case in which no reduction process is applied---denoted as ALL---are included. Also, let subscript $m$ represent the input parameter of the PG methods when required, \textit{i.e.} MChen$_{m}$, MRSP1$_{m}$, and MRSP2$_{m}$, which relates to the number of partitions as $n_{d} = m\cdot|\mathcal{T}_{ml}|/100$. For assessing its influence in the scheme, we considered different values of this input parameter as $m\in\left\{10, 30, 50, 70, 90\right\}$. 

The remainder of the section presents four particular experiments: (i) a first part in which the PG methods are comparatively evaluated obviating the noise induction process; (ii) a second one whose focus is the noise robustness and data cleansing capabilities of these PG schemes; (iii) a third passage that assesses the PG methods from the perspective of class-imbalance data; and (iv) a last part that benchmarks these strategies in terms of their execution time.

The implementation of the proposed PG methods and the experimental procedure considered is publicly available in: \url{https://github.com/jose-jvmas/multilabel_PG}. In addition, all obtained results for each individual corpus, configuration, and scenario contemplated are available at Mendeley Data (\url{http://dx.doi.org/10.17632/rbcnc6jcf3}) for every single experiment performed in the work.

\subsection{Comparative assessment of multilabel PG strategies}\label{subsec:ComparativeAssessment}

In this first experiment, we thoroughly compare the different reduction strategies using the aforementioned multilabel $k$NN-based classifiers as individual scenarios. In this regard, Table~\ref{tab:Results_Individual_Classifiers} and Figure~\ref{fig:Results_NoNoise} show the results obtained in which the performance and reduction figures constitute the average of the individual values obtained for the corpora considered. 

\begin{table}[!ht]
    \centering
    \caption{Results in terms of HL and resulting size for both the reference methods (exhaustive search, denoted as ALL, and MRHC) and our proposals (MChen, MRSP1, MRSP2, and MRSP3) when considering the different $k$NN-based classifiers. Non-dominated solutions per classifier are highlighted in bold type. Underlined values denote the best performance rates per PG scheme and classifier.}
    \label{tab:Results_Individual_Classifiers}
    \setlength{\tabcolsep}{3.5pt}
    \renewcommand{\arraystretch}{.8}
    \resizebox{\textwidth}{!}{%
    \begin{tabular}{llcccccccccccccccc}
        \toprule[1pt]
        & & \multirow{2}{*}{\textbf{Size}} && \multicolumn{4}{c}{\textbf{BR\textit{k}NN}} && \multicolumn{4}{c}{\textbf{LP-\textit{k}NN}} && \multicolumn{4}{c}{\textbf{ML-\textit{k}NN}}\\
        \cmidrule(lr){5-8} \cmidrule(lr){10-13} \cmidrule(lr){14-18}
        & & && 1 & 3 & 5 & 7 && 1 & 3 & 5 & 7 && 1 & 3 & 5 & 7\\
        \cmidrule(lr){1-18}
        \multicolumn{4}{l}{\textbf{Reference}}\\
        & ALL \small{$\blacksquare$} & 100 && 9.09 & 7.94 & 7.69 & \underline{7.56} && 9.09 & 8.71 & 8.54 & \underline{8.45} && 9.09 & 7.92 & 7.72 & \underline{7.66}\\
        & MRHC \textcolor{MRHC_color}{$\blacktriangledown$} & 59.62  && 8.76 & 7.70 & \underline{7.49} & 7.51 && 8.76 & \underline{8.47} & 8.57 & 8.68 && 8.76 & 7.87 & 7.91 & \underline{7.85}\\
        \multicolumn{2}{l}{\textbf{Proposals}}\\
        & MChen$_{10}$ \symMChen & \textbf{\phantom{0}9.98}   && 7.92 & \underline{\textbf{7.74}} & 7.78 & 7.83 && 7.92 & \underline{\textbf{7.90}} & 7.98 & 7.97 && 7.92 & 7.87 & \underline{\textbf{7.75}} & 7.86\\
        & MChen$_{30}$ \symMChen & \textbf{29.94}  && 8.00 & 7.58 & \underline{\textbf{7.51}} & 7.55 && 8.00 & 7.91 & \underline{\textbf{7.86}} & 7.87 && 8.00 & \underline{7.70} & 7.77 & 7.81\\
        & MChen$_{50}$ \symMChen & 49.96  && 8.29 & 7.71 & 7.53 & \underline{7.38} && 8.29 & 8.30 & \underline{8.06} & 8.10 && 8.29 & 7.85 & 7.63 & \underline{7.57}\\
        & MChen$_{70}$ \symMChen & 69.97  && 8.51 & 7.72 & 7.62 & \underline{7.46} && 8.51 & \underline{8.35} & 8.40 & 8.49 && 8.51 & 7.90 & 7.82 & \underline{\textbf{7.45}}\\
        & MChen$_{90}$ \symMChen & 89.02  && 8.73 & 7.62 & 7.29 & \underline{7.17} && 8.73 & 8.35 & \underline{8.16} & 8.33 && 8.73 & 7.61 & \underline{7.53} & 7.55\\
        \cdashline{2-18}
        & MRSP1$_{10}$ \symMRSPOne & 61.88  && 8.95 & 7.96 & 7.60 & \underline{7.40} && 8.95 & \underline{8.88} & 9.03 & 8.97 && 8.95 & 8.21 & 7.95 & \underline{7.68}\\
        & MRSP1$_{30}$ \symMRSPOne & \textbf{74.51}  && 8.77 & 7.76 & 7.36 & \underline{\textbf{7.17}} && 8.77 & 8.55 & \underline{8.34} & 8.50 && 8.77 & 7.84 & \underline{\textbf{7.44}} & 7.56\\
        & MRSP1$_{50}$ \symMRSPOne & 80.11  && 8.80 & 7.78 & 7.44 & \underline{7.27} && 8.80 & 8.41 & \underline{8.34} & 8.50 && 8.80 & 7.84 & \underline{7.50} & 7.54\\
        & MRSP1$_{70}$ \symMRSPOne & 84.37  && 8.86 & 7.77 & 7.45 & \underline{7.30} && 8.86 & 8.43 & \underline{8.32} & 8.46 && 8.86 & 7.79 & 7.54 & \underline{7.52}\\
        & MRSP1$_{90}$ \symMRSPOne & 90.78  && 8.84 & 7.71 & 7.33 & \underline{7.17} && 8.84 & 8.46 & \underline{8.27} & 8.46 && 8.84 & 7.68 & \underline{7.51} & 7.57\\
        \cdashline{2-18}
        & MRSP2$_{10}$ \symMRSPTwo & 61.09  && 8.85 & 7.90 & 7.61 & \underline{7.36} && 8.85 & 8.92 & 8.93 & \underline{8.87} && 8.85 & 8.04 & 7.86 & \underline{7.79}\\
        & MRSP2$_{30}$ \symMRSPTwo & 78.07  && 8.79 & 7.87 & 7.53 & \underline{7.34} && 8.79 & 8.52 & \underline{8.26} & 8.39 && 8.79 & 7.88 & 7.68 & \underline{7.47}\\
        & MRSP2$_{50}$ \symMRSPTwo & \textbf{83.59}  && 8.74 & 7.76 & 7.45 & \underline{7.27} && 8.74 & 8.59 & 8.36 & \underline{8.34} && 8.74 & 7.79 & 7.58 & \underline{\textbf{7.43}}\\
        & MRSP2$_{70}$ \symMRSPTwo & \textbf{87.21}  && 8.80 & 7.65 & 7.38 & \underline{\textbf{7.12}} && 8.80 & 8.44 & 8.36 & \underline{8.30} && 8.80 & 7.69 & 7.66 & \underline{7.52}\\
        & MRSP2$_{90}$ \symMRSPTwo & 89.28  && 8.76 & 7.68 & 7.49 & \underline{7.34} && 8.76 & 8.47 & 8.51 & \underline{8.34} && 8.76 & 7.80 & 7.51 & \underline{7.50}\\
        \cdashline{2-18}
        & MRSP3 \symMRSPThree & 66.88  && 8.53 & 7.62 & 7.50 & \underline{7.34} && 8.53 & 8.29 & \underline{8.12} & 8.23 && 8.53 & 7.94 & 7.78 & \underline{7.71}\\
        \bottomrule[1pt]
    \end{tabular}
    }
\end{table}

\begin{figure}[!ht]
    \centering
    \includegraphics[width=.875\textwidth]{./figures/Results_NoNoise.pdf}
    \caption{Results in terms of HL and resulting size obtained with the $k$NN-based classifiers when considering the PG methods and the exhaustive search case (ALL) for the different $k$ values tested. Circled methods and dashed lines represent the non-dominated elements and the Pareto frontiers in each scenario, respectively. For easier comparison, shaded areas depict the regions in the solution space occupied by the baseline cases (MRHC and ALL).}
    \label{fig:Results_NoNoise}
\end{figure}

A first remark that may be observed is that the proposed methods fill a region in the space of possible solutions not previously occupied by existing multilabel PG methods. This is because some of the proposals (MChen, MRSP1, and MRSP2) allow selecting the size of the reduced set through a parameter. Note that, while this may be considered a drawback, such a feature allows prioritising either the reduction rate or classification performance depending on the particular application considered.

It can be also checked that, for all cases, MChen achieves the highest reduction rates, even when other parameter-based multilabel PG proposals consider the same $m$ value. The main reason for such an effect is that, for a given $m$ reduction setting, the MChen performs a rather aggressive reduction---especially with low $m$ values---as it only retrieves a single prototype per region. On the contrary, the merging procedure for both MRSP1 and MRSP2 softens the single-prototype policy by the MChen proposal to compute an instance per existing labelset in the partitions, hence increasing the size of the $R_{ml}$ resulting set. Regarding the MRSP3 method, the fact that the resulting set size may be deemed as medium-to-high (a $66.88\%$ of the size of the ALL case) is due to the region-based homogeneity requirement of the space partitioning phase; in this sense, a more relaxed criterion (\textit{e.g.}, allowing only partial label matches) should result in sharper reduction rates.

In all cases, since the PG process is applied before the classification stage, the resulting set sizes are the same for all scenarios, being the differences in performance only due to the particular capabilities of the classification scheme. It can be observed that LP-$k$NN may be deemed as the least competitive alternative since, for the same reduction scheme, HL figures tend to be higher than the other alternatives. Oppositely, BR$k$NN and ML-$k$NN show similar performance results since the HL figures do not remarkably differ among them. Such performance disparities among the classifiers are most likely due to the restrictiveness of the LP-$k$NN classifier that, in contrast to the BR$k$NN and ML-$k$NN methods, is not able to infer labelsets not seen during the training stage.

From the point of view of the PG strategies, the rather sharp reduction figures depicted by the MChen method---mainly due to the single-prototype policy of the merging stage---generally entails the least competitive classification rates among the PG techniques, being the sole exception found when contemplating the LP-$k$NN classifier. This fact is most probably due to that the resulting prototypes in that scenario only comprise the most relevant labels in the corpora, being hence guaranteed the inference of a representative part of the classes at the expense of missing sporadic labels. The rest of the cases---MRHC and the entire MRSP family---generally achieve better classification rates than the MChen alternative---and especially the MRSP1 and MRSP2 methods---for the different $k$NN-based classifiers. Again, an exception is found when considering the LP-$k$NN one, which is most likely due to the high label variability in the search space that visibly hinders the performance of this classifier.

In terms of non-dominance, it may be noted that the obtained Pareto frontiers in the different classification scenarios considered only comprise examples of the novel multilabel PG strategies proposed in the work: BR$k$NN contains MChen$_{10}$, MChen$_{30}$, MRSP1$_{30}$, and MRSP2$_{70}$; LP-$k$NN depicts the MChen$_{10}$ and MChen$_{30}$ cases; and ML-$k$NN points out four of them, which are MChen$_{10}$, MChen$_{70}$, MRSP1$_{30}$, and MRSP2$_{90}$. Hence, the ALL and MRHC cases may not be considered optimal solutions to the task as they are consistently dominated by the novel proposals presented in this work.

Finally, it may be also checked that the classification rates do generally improve as the number of neighbours considered---$k$ parameter of the classifiers---increases. This fact suggests the presence of some noise in the corpora that is somehow palliated by adequately tuning this parameter. Note that, among the different multilabel classifiers studied, LP-$k$NN is the one that shows the least improvement when increasing this $k$ value.

\subsubsection{Statistical significance analysis}

A significance analysis has been performed to statistically evaluate the results obtained. For that, we have considered the Wilcoxon signed-rank test~\citep{demvsar2006statistical} to assess whether the classification performance and reduction rate of the proposed PG methods significantly improve those of the baseline strategies. More precisely, for each classification scenario, we compare the results obtained by the elements of the particular Pareto frontier against the best figures obtained by the baseline MRHC and ALL methods. For that, we consider the individual results obtained---either performance or reduction---for each contemplated corpus in the experimentation. Table~\ref{tab:results_PG_AllCases_Wilcoxon} shows the results of such analysis when considering a significance threshold of $p < 0.05$. 

\begin{table}[!ht]
\caption{Wilcoxon signed-rank test results with $p < 0.05$ for the classifiers considered. Symbols \statV, \statX, and $=$ respectively denote that, for each classification scenario, the non-dominated solution in the row significantly improves, worsens or does not differ from the reference one in the column for the performance (HL) or reduction (Size) criterion.}
\label{tab:results_PG_AllCases_Wilcoxon}
\centering
\renewcommand{\arraystretch}{.7}
% \resizebox{\textwidth}{!}{%
\begin{tabular}{llcccccc}
\toprule[1pt]
 &&&  \multicolumn{2}{c}{\textbf{ALL} \small{$\blacksquare$}} && \multicolumn{2}{c}{\textbf{MRHC} \textcolor{MRHC_color}{$\blacktriangledown$}}\\
 \cmidrule(lr){4-5} \cmidrule(lr){7-8}
 &&& HL & Size && HL & Size\\
\cmidrule(lr){1-8}
\multicolumn{2}{l}{\textbf{BR\textit{k}NN}}\\
& MChen$_{10}$ \symMChen && = & \statV && = & \statV\\
& MChen$_{30}$ \symMChen && = & \statV && = & \statV\\
& MRSP1$_{30}$ \symMRSPOne && \statV & \statV && \statV & =\\
& MRSP2$_{70}$ \symMRSPTwo && \statV & \statV && \statV & \statX\\
\cmidrule(lr){1-8}
\multicolumn{2}{l}{\textbf{LP-\textit{k}NN}}\\
& MChen$_{10}$ \symMChen && = & \statV && = & \statV\\
& MChen$_{30}$ \symMChen && \statV & \statV && \statV & \statV\\
\cmidrule(lr){1-8}
\multicolumn{2}{l}{\textbf{ML-\textit{k}NN}}\\
& MChen$_{10}$ \symMChen && = & \statV && = & \statV\\
& MChen$_{70}$ \symMChen && = & \statV && \statV & =\\
& MRSP1$_{30}$ \symMRSPOne && = & \statV && = & =\\
& MRSP2$_{50}$ \symMRSPTwo && = & \statV && \statV & \statX\\
 \bottomrule[1pt]
\end{tabular}
% }
\end{table}

Focusing on the classification performance criterion (HL), it may be observed that the non-dominated elements in the Pareto frontier---exclusively defined by the proposals introduced in the work---statistically equal or improve the results of the baselines considered. However, since the particular conclusions are quite related to the actual classification scheme at hand, we shall now analyse them in a separate manner.

When considering the BR$k$NN classifier, the proposals depicting the highest reduction rates---MChen$_{10}$ and MChen$_{30}$---show similar performance to both ALL and MRHC baseline cases; on the contrary, those schemes with larger resulting set sizes---MRSP1$_{30}$ and MRSP2$_{70}$---do improve the reference strategies.

In the case of the LP-$k$NN classifier, a similar trend to that of the BR$k$NN is found: when performing a sharp reduction---MChen$_{10}$ strategy---, the reported classification rate does not statistically differ to those of the baselines; however, when allowing a larger set size---the MChen$_{30}$ method---, this performance indicator does improve those of the reference cases.

The results obtained with the ML-$k$NN classifier, however, do not show a similar tendency to the ones presented. As it may be observed, none of the non-dominated cases is able to statistically outperform the ALL case, while they do obtain similar performance scores with remarkably fewer prototypes. Regarding the MRHC base case, two of the proposals---MChen$_{70}$ and MRSP2$_{50}$---do significantly improve this base case while the other two non-dominated elements---MChen$_{10}$ and MRSP1$_{30}$---report statistically similar classification rates.

In relation to the analysis of the reduction capabilities, as expected, the results show that all non-dominated cases statistically improve the ALL case. Oppositely, when compared to the MRHC, there is a larger variability in the results: MChen generally outperforms the reference method except for the case of MChen$_{70}$ in the ML-$k$NN scenario, which shows no statistical difference; the MRSP1$_{30}$---found in BR$k$NN and ML-$k$NN---also shows alike reduction capabilities to MRHC as the analysis points out no difference; finally, MRSP2$_{70}$ and MRSP2$_{50}$, respectively found in the BR$k$NN and ML-$k$NN scenarios, stand for the cases in which the reduction results are statistically worse than MRHC, given that these methods do not remarkably reduce the set size of the reference corpus.

\subsection{Noise robustness and data cleansing study}

In this second experiment, we assess the performance of both the proposed multilabel PG strategies as well as the reference ones in scenarios with noisy data. For that, we consider the labelset swapping procedure introduced in Section~\ref{subsect:noise_induction} with $\theta\in\left\{20\%,40\%\right\}$ as they stand as representative noise rates commonly considered in the related literature~\cite{natarajan2013learning}. For comparative purposes, the case of $\theta=0$ is also included to assess the base case in which no noise is induced. Note that, the different corpora in each experiment are affected by the same level of induced noise---\textit{i.e.,} the same $\theta$ value for all corpora---, being the case of different noise levels per corpus posed as future work.

\begin{figure}[!ht]
    \centering
    \includegraphics[width=.875\textwidth]{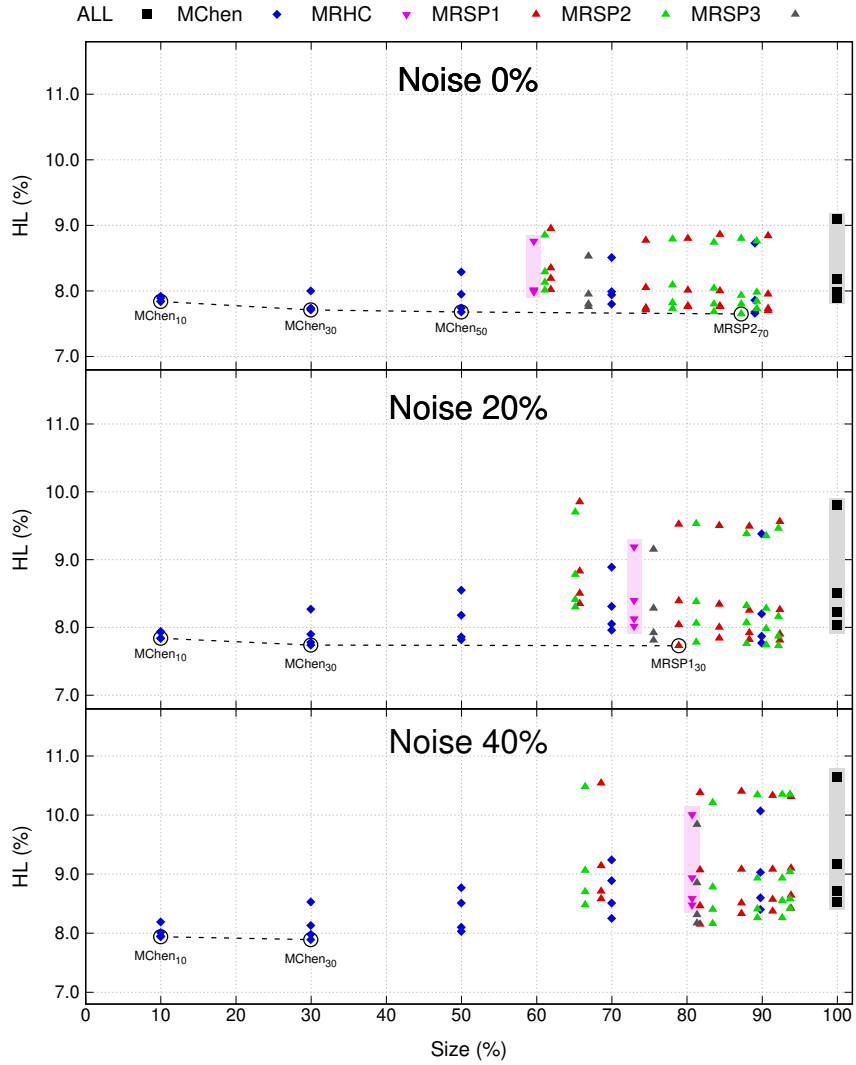}
    \caption{Results in terms of HL and resulting size for the different noise scenarios when considering the PG methods and the exhaustive search case (ALL) for the $k$ values tested. Note that each sample constitutes the average performance obtained for the three classification methods studied. Circled methods and dashed lines represent the non-dominated elements and the Pareto frontiers in each scenario, respectively. For easier comparison, shaded areas depict the regions in the solution space occupied by the baseline cases (MRHC and ALL).}
    \label{fig:Results_Noise}
\end{figure}

The results obtained in the different noise scenarios posed are depicted in Table~\ref{tab:Results_Noise_Induction} and Figure~\ref{fig:Results_Noise}. Note that, for simplicity, these figures constitute the average performance---both in terms of recognition rate and reduction capabilities---of the individual results per PG method and $k$ classification parameter for the three $k$NN-based algorithms considered.

\begin{table}[!ht]
    \centering
    \caption{Results in terms of HL and resulting size for both the reference methods (exhaustive search, denoted as ALL, and MRHC) and our proposals (MChen, MRSP1, MRSP2, and MRSP3) when considering the different noise scenarios posed. Each value constitutes the average performance obtained for the three classification methods considered. Non-dominated solutions per noise scenario are highlighted in bold type. Underlined values denote the best performance rates per PG scheme and noise scenario.}
    \label{tab:Results_Noise_Induction}
    \setlength{\tabcolsep}{3.5pt}
    \renewcommand{\arraystretch}{.8}
    \resizebox{\textwidth}{!}{%
    \begin{tabular}{llccccccccccccccccc}
        \toprule[1pt]
        & & \multicolumn{5}{c}{\textbf{Noise 0\%}} && \multicolumn{5}{c}{\textbf{Noise 20\%}} && \multicolumn{5}{c}{\textbf{Noise 40\%}}\\
        \cmidrule(lr){3-7} \cmidrule(lr){9-13} \cmidrule(lr){15-19}
        & & \multirow{2}{*}{Size} & \multicolumn{4}{c}{k} && \multirow{2}{*}{Size} &  \multicolumn{4}{c}{k} && \multirow{2}{*}{Size} &   \multicolumn{4}{c}{k}  \\
        \cmidrule(lr){4-7} \cmidrule(lr){10-13} \cmidrule(lr){16-19}
        & &  & 1 & 3 & 5 & 7 && & 1 & 3 & 5 & 7 && & 1 & 3 & 5 & 7\\
        \cmidrule(lr){1-19}
        \multicolumn{2}{l}{\textbf{Reference}}\\
        & ALL \small{$\blacksquare$} & 100 & 9.09 & 8.19 & 7.98 & \underline{7.89} && 100 & 9.80 & 8.50 & 8.22 & \underline{8.03} && 100 & 10.65 & 9.17 & 8.71 & \underline{8.53}\\
	& MRHC \textcolor{MRHC_color}{$\blacktriangledown$} & 59.62   &   8.76  &    8.01 &  \underline{7.99}    & 8.02   && 72.95 &  9.19    & 8.40   & 8.13  &  \underline{8.02} &&    80.67   &   10.01 &    8.94    &   8.59   &   \underline{8.48}\\
	\multicolumn{2}{l}{\textbf{Proposals}}\\
	& MChen$_{10}$ \symMChen & \textbf{\phantom{0}9.98}    &   7.92   &   \underline{\textbf{7.84}}  &    \underline{7.84} &  7.89    && \textbf{\phantom{0}9.98}   & 7.94  &  \underline{\textbf{7.84}} &    \underline{7.84}    &   7.93   &&   \textbf{\phantom{0}9.98}  &    8.19 &  7.95    & \underline{\textbf{7.94}}   & 8.01\\
	& MChen$_{30}$ \symMChen & \textbf{29.94}   &   8.00  &    7.73 &  \underline{\textbf{7.71}}    & 7.74   && \textbf{29.94} &  8.27    & 7.90   & \underline{\textbf{7.74}}  &  7.78 &&    \textbf{29.94}   &   8.53  &    8.13 &  \underline{\textbf{7.89}}    & 7.98\\
	& MChen$_{50}$ \symMChen & \textbf{49.96}   &   8.29  &    7.95 &  7.74    & \underline{\textbf{7.68}}   && 49.96 &  8.55    & 8.18   & 7.86  &  \underline{7.82} &&    49.96   &   8.77  &    8.51 &  8.10    & \underline{8.03}\\
	& MChen$_{70}$ \symMChen & 69.97   &   8.51  &    7.99 &  7.94    & \underline{7.80}   && 69.97 &  8.89    & 8.31   & 8.05  &  \underline{7.96} &&    69.97   &   9.24  &    8.89 &  8.51    & \underline{8.25}\\
	& MChen$_{90}$ \symMChen & 89.02   &   8.73  &    7.86 &  \underline{7.66}    & 7.68   && 89.02 &  9.38    & 8.20   & 7.87  &  \underline{7.77} &&    89.02   &   10.07 &    9.03    &   8.60   &   \underline{8.40}\\
	\cdashline{2-19}
	& MRSP1$_{10}$ \symMRSPOne & 61.88   &   8.95  &    8.35 &  8.19    & \underline{8.02}   && 65.72 &  9.85    & 8.83   & 8.50  &  \underline{8.35} &&    68.56   &   10.54 &    9.14    &   8.71   &   \underline{8.58}\\
	& MRSP1$_{30}$ \symMRSPOne & 74.51   &   8.77  &    8.05 &  \underline{7.71}    & 7.74   && \textbf{78.89} &  9.52    & 8.39   & 8.04  &  \underline{\textbf{7.73}} &&    81.73   &   10.38 &    9.07    &   8.46   &   \underline{8.15}\\
	& MRSP1$_{50}$ \symMRSPOne & 80.11   &   8.80  &    8.01 &  \underline{7.76}    & 7.77   && 84.29 &  9.50    & 8.34   & \underline{8.00}  &  7.84 &&    87.23   &   10.40 &    9.08    &   8.51   &   \underline{8.33}\\
	& MRSP1$_{70}$ \symMRSPOne & 84.37   &   8.86  &    8.00 &  7.77    & \underline{7.76}   && 88.28 &  9.49    & 8.25   & 7.92  &  \underline{7.82} &&    91.38   &   10.33 &    9.08    &   8.57   &   \underline{8.37}\\
	& MRSP1$_{90}$ \symMRSPOne & 90.78   &   8.84  &    7.95 &  \underline{7.70}    & 7.73   && 92.35 &  9.56    & 8.26   & 7.90  &  \underline{7.81} &&    93.82   &   10.31 &    9.10    &   8.64   &   \underline{8.42}\\
	\cdashline{2-19}
	& MRSP2$_{10}$ \symMRSPTwo & 61.09   &   8.85  &    8.29 &  8.13    & \underline{8.01}   && 65.13 &  9.70    & 8.78   & 8.41  &  \underline{8.30} &&    66.44   &   10.48 &    9.06    &   8.70   &   \underline{8.48}\\
	& MRSP2$_{30}$ \symMRSPTwo & 78.07   &   8.79  &    8.09 &  7.82    & \underline{7.73}   && 81.23 &  9.53    & 8.38   & 8.06  &  \underline{7.78} &&    83.41   &   10.21 &    8.78    &   8.40   &   \underline{8.16}\\
	& MRSP2$_{50}$ \symMRSPTwo & 83.59   &   8.74  &    8.04 &  7.80    & \underline{7.68}   && 87.92 &  9.38    & 8.32   & 8.07  &  \underline{7.76} &&    89.37   &   10.34 &    8.93    &   8.41   &   \underline{8.26}\\
	& MRSP2$_{70}$ \symMRSPTwo & \textbf{87.21}   &   8.80  &    7.93 &  7.80    & \underline{\textbf{7.65}}   && 90.53 &  9.35    & 8.28   & \underline{7.98}  &  7.74 &&    92.67   &   10.35 &    8.93    &   8.55   &   \underline{8.26}\\
	& MRSP2$_{90}$ \symMRSPTwo & 89.28   &   8.76  &    7.98 &  7.84    & \underline{7.73}   && 92.16 &  9.46    & 8.16   & 7.87  &  \underline{7.73} &&    93.69   &   10.35 &    9.04    &   8.58   &   \underline{8.42}\\
	\cdashline{2-19}
	& MRSP3 \symMRSPThree & 66.88   &   8.53  &    7.95 &  7.80    & \underline{7.76}   && 75.54 &  9.15    & 8.28   & 7.92  &  \underline{7.81} &&    81.33   &   9.84  &    8.85 &  8.31    & \underline{8.17}\\
        \bottomrule[1pt]
    \end{tabular}
    }
\end{table}

The induction of noise in the corpora clearly affects the overall performance since, in general, all studied cases depict lower classification rates as the noise level increases. While the use of high $k$ classification values (e.g., $k=5$ or $k=7$) somehow palliates this effect, the best performance achieved in these noisy scenarios is indeed lower than that of the non-induced noise case.

Besides, all PG methods generally show worse reduction rates as the noise increases, being the MRHC and MRSP3 strategies particularly affected. Most likely, the induced noise results in a higher labelset diversity---\textit{i.e.}, less label-level homogeneity---in the prototype groups obtained during the space partitioning stage of the methods, hence forcing the merging stage to generate a larger number of elements to satisfy the condition of retrieving as many prototypes as labelsets in the cluster. The sole exception to this assertion is the MChen method whose reduction capabilities remain stable independently of the noise induced in the data since the prototype merging policy of this strategy always retrieves a single instance per partition.

Overall, it can be noted that MChen can be considered the best noise cleansing strategy since the classification schemes trained after that stage achieve the best overall HL performance figures. Most reasonably, since the merging policy of this strategy only keeps the most common labels in each cluster retrieved from the space partitioning stage, the method inherently removes the sporadic presence of uncommon labels in each of these groups, thus showing the aforementioned noise robustness. MRSP proposals, though, prove not to be that competitive against this type of noise since the performance of the classification schemes trained with the set obtained with those methods degrades as the presence of noise increases. Note that, since these merging policies produce as many prototypes as label combinations exist in each data partition, the higher labelset variability due to the noise induction process not only results in the generation of a larger amount of prototypes, but also that they may be incorrectly labelled. Regarding the reference MRHC method, it may be observed a similar performance trend to that of the MRSP family as they are based on similar reduction principles.

In terms of non-dominance, it may be observed that the different Pareto frontiers are entirely defined by the novel PG proposals introduced in this work: MChen$_{10}$, MChen$_{30}$, MChen$_{50}$, and MRSP2$_{70}$ in the non-induced noise scenario; MChen$_{10}$, MChen$_{30}$, and MRSP1$_{30}$ when $\theta=20\%$; and MChen$_{10}$ and MChen$_{30}$ when considering the noisiest scenario of the ones studied in the work. In this regard, it may be concluded that the MChen algorithm proves itself as a considerably robust method---both in terms of efficiency and classification performance---against noisy situations, especially when set to high reduction rates (e.g., $m=10\%$ or $m=30\%$).

It must be noted that DR methods based on editing strategies are typically contemplated in multiclass scenarios as a means of removing noisy elements from the data to enhance the performance of a subsequent DR or classification technique~\cite{garcia2015data}. In this context, and according to their reported noise removal capabilities, multilabel editing strategies such as the ones by Kanj et al.~\cite{KanjAbdallahDenoeuxTout:TPAMI:2016} or Arnaiz-Gonz{\'a}lez et al.~\cite{ArnaizGonzalezDiezPastorRodriguezGarciaOsorio:ASOC:2018} may be used for performing such a noise cleansing process before applying any particular multilabel PG method.

\subsubsection{Statistical significance analysis}

As in the first experiment, we have considered the Wilcoxon signed-rank test to statistically compare the results obtained by the elements of the Pareto frontier against the best results obtained by the baseline MRHC and ALL methods for each noise scenario. Table~\ref{tab:results_PG_Noise_Wilcoxon} shows the outcome of such analysis when considering a significance threshold of $p < 0.05$.

\begin{table}[!ht]
\caption{Wilcoxon signed-rank test results with $p < 0.05$ for the noise scenarios posed. Symbols \statV, \statX, and $=$ respectively denote that, for each classification scenario, the non-dominated solution in the row significantly improves, worsens or does not differ from the reference one in the column for the performance (HL) or reduction (Size) criterion.}
\label{tab:results_PG_Noise_Wilcoxon}
\centering
\renewcommand{\arraystretch}{.7}
% \resizebox{\textwidth}{!}{%
\begin{tabular}{llcccccc}
\toprule[1pt]
 &&&  \multicolumn{2}{c}{\textbf{ALL} \small{$\blacksquare$}} && \multicolumn{2}{c}{\textbf{MRHC} \textcolor{MRHC_color}{$\blacktriangledown$}}\\
 \cmidrule(lr){4-5} \cmidrule(lr){7-8}
 &&& HL & Size && HL & Size\\
\cmidrule(lr){1-8}
\multicolumn{2}{l}{\textbf{Noise 0\%}}\\
& MChen$_{10}$ \symMChen && = & \statV && = & \statV\\
& MChen$_{30}$ \symMChen && = & \statV && \statV & \statV\\
& MChen$_{50}$ \symMChen && = & \statV && \statV & \statV\\
& MRSP2$_{70}$ \symMRSPTwo && = & \statV && \statV & \statV\\
\cmidrule(lr){1-8}
\multicolumn{2}{l}{\textbf{Noise 20\%}}\\
& MChen$_{10}$ \symMChen && = & \statV && \statV & \statV\\
& MChen$_{30}$ \symMChen && = & \statV && \statV & \statV\\
& MRSP1$_{30}$ \symMRSPOne && \statV & \statV && \statV & \statV\\
\cmidrule(lr){1-8}
\multicolumn{2}{l}{\textbf{Noise 40\%}}\\
& MChen$_{10}$ \symMChen && \statV & \statV && \statV & \statV\\
& MChen$_{30}$ \symMChen && \statV & \statV && \statV & \statV\\
 \bottomrule[1pt]
\end{tabular}
% }
\end{table}

As it may be observed, the multilabel PG strategies proposed in the work significantly improve the reduction rate of the baselines considered for all noise scenarios posed. Such a point suggests a remarkable robustness of our methods to the presence of noise in the data: while the reduction capabilities of the reference MRHC strategy severely degrade as the noise in the data increases, the MChen is not affected by such an alteration whereas the MRSP1 and MRSP2 strategies do not degrade as much as the MRHC.

In relation to the classification rate, it may be noted that all non-dominated proposals either equal or improve the exhaustive search case with a significantly lower amount of prototypes. More precisely, our proposals improve the ALL case when inducing an elevated level of noise in the data while, when addressing scenarios with low levels of induced noise, the proposed multilabel methods in the Pareto frontier do not significantly differ from the exhaustive search cases.

Regarding the classification performance of the MRHC baseline method, it may be observed that this strategy is remarkably affected by the noise, being significantly outperformed by all the multilabel PG proposals in the non-dominated frontier. The sole exception to this assertion is the MChen$_{10}$ in the Noise 0\% scenario that does not significantly differ from the MRHC case.

Overall, this analysis proves the superior robustness and noise cleansing capabilities of the proposed multilabel PG alternatives since, in the worst-case scenario, the classification rate achieved is similar to that of the exhaustive search but with a significantly lower amount of samples. Besides, it is also proved that the only existing multilabel PG method in the literature---the MRHC algorithm---is severely affected by these noisy scenarios---both in terms of efficiency and classification rate---, being hence outperformed by the novel multilabel PG proposals introduced in this work.

%-------------------------------------------------------------------------------------------------
\subsection{Class imbalance analysis}

In order to provide some additional insights regarding the capabilities of the multilabel PG methods, this third experiment assesses their performance by attending to the label-based imbalance ratio of the different corpora used in the work. More precisely, to observe possible relations between the imbalance degree and the overall performance, we have gathered the different corpora based on their respective \textit{MeanIR} score (see Table~\ref{tab:corpora}): a first moderate class imbalance group ($10\leq\mbox{\textit{MeanIR}}<100$) comprising the \textit{bibtex}, \textit{genbase}, and \textit{medical} sets; and a second highly-imbalanced collection with ($\mbox{\textit{MeanIR}}\geq100$) containing \textit{Corel5k} and all the \textit{rcv1subset} corpora.

Regarding the evaluation procedures, two representative example-based figures of merit for imbalance data have been considered~\cite{LiuBlekasTsoumakas:PR:2022}: the F-measure (F$^{S}_{1}$) and the Area Under the Receiver Operating Characteristic Curve (AUC$^{S}$). Based on the notation introduced in this work, these metrics are defined as:
\begin{align} 
    \mbox{F}_{1}^{S} &= \frac{1}{\left|\mathcal{S}_{ml}\right|}\cdot\sum_{i=1}^{\left|\mathcal{S}_{ml}\right|}\frac{2\cdot|\mathbcal{y}_{i} \cap \hat{h}_{ml}\left(\mathbcal{x}_{i}\right)|}{|\mathbcal{y}_{i}| + |\hat{h}_{ml}\left(\mathbcal{x}_{i}\right)|}\\[7pt]
    \mbox{AUC}^{S} &= \frac{1}{\left|\mathcal{S}_{ml}\right|}\cdot\sum_{i=1}^{\left|\mathcal{S}_{ml}\right|}\frac{|\hat{h}_{ml}\left(\mathbcal{x}_{i}\right)|}{|\mathbcal{y}_{i}| \cdot \left(\left|\mathcal{Y}_{ml}\right|-1\right)}
\end{align}
\noindent where $\mathcal{S}_{ml} \subset \mathcal{X}\times\mathcal{Y}_{ml}$ denotes the multilabel set of test data, elements $\mathbcal{y}_{i}$ and $\hat{h}_{ml}\left(\bm{x}_{i}\right)$ respectively stand for the ground-truth and predicted labelsets, and $\mathcal{Y}_{ml}$ denotes the target label space.

Considering all the above, Table~\ref{tab:imbalanced_metrics} presents the results obtained for the aforementioned imbalance-based corpora assortments---namely, \textit{Moderate} and \textit{High}---together with the case of examining all data collections---denoted as \textit{All corpora}---for the two contemplated metrics as well as their average resulting size. Note that, for the sake of conciseness and comparison with the analyses performed in the previous sections, this study assesses the non-dominated solutions per classification scheme obtained in Section~\ref{subsec:ComparativeAssessment} as well as the best-performing configurations for the baseline cases.

\begin{table}[!ht]
\caption{Results in terms of the F$^{S}_{1}$ (\%) and AUC$^{S}$ (\%) figures of merit for the two class imbalance ranges considered---denoted as \textit{Moderate} and \textit{High}---as well as for all the entire data collection---designated as \textit{All corpora}---for the non-dominated solutions in Section~\ref{subsec:ComparativeAssessment} per classification scenario. Figures reported for the reference ALL and MRHC strategies constitute those obtained with the best performing $k$ classification parameter for each particular case. The best performance figures per classifier, imbalance level, and metric are highlighted in bold. The resulting set size is provided for comparative purposes.}
\label{tab:imbalanced_metrics}
\centering
\renewcommand{\arraystretch}{.8}
\resizebox{\textwidth}{!}{%
\begin{tabular}{llcccccccccccc}
\toprule[1pt]
 &&&  \multicolumn{3}{c}{\textbf{Moderate}} & \multicolumn{3}{c}{\textbf{High}}  & \multicolumn{3}{c}{\textbf{All corpora}}\\
 \cmidrule(lr){4-6} \cmidrule(lr){7-9} \cmidrule(lr){10-12}
 &&& F$^{S}_{1}$ & AUC$^{S}$ & Size & F$^{S}_{1}$ & AUC$^{S}$ &  Size & F$^{S}_{1}$ & AUC$^{S}$ &  Size \\
 \cmidrule(lr){1-12}
\multicolumn{2}{l}{\textbf{BR\textit{k}NN}}\\
& ALL \small{$\blacksquare$} && 23.09 & 60.88 & 100 & \textbf{3.50} & \textbf{50.95} & 100 & 21.55 & 58.95 & 100\\
& MRHC \textcolor{MRHC_color}{$\blacktriangledown$} && 22.87 & 60.60 & 60.80 & 3.02 & 50.83 & 55.55 & 20.33 & 58.43 & 59.62\\
\cdashline{2-12}
& MChen$_{10}$ \symMChen && 14.35 & 56.44 & \phantom{0}9.95 & 1.25 & 50.16 & 10.00 & 14.01 & 55.60 &	\phantom{0}9.98\\
& MChen$_{30}$ \symMChen && 11.77 & 55.20 & 29.85 & 1.44 & 50.21 & 30.00 & 15.03 & 56.00 & 29.94\\
& MRSP1$_{30}$ \symMRSPOne && 20.66 & 59.77 & 75.44 & 2.81 & 50.82 & 80.86 & 20.70 & 58.72 & 74.51\\
& MRSP2$_{70}$ \symMRSPTwo && \textbf{24.10} & \textbf{61.47} & 88.35 & 2.49 & 50.75 & 87.63 & \textbf{21.71} & \textbf{59.21} & 87.21\\
\cmidrule(lr){1-12}
\multicolumn{2}{l}{\textbf{LP-\textit{k}NN}}\\
& ALL \small{$\blacksquare$} && 35.29 & 66.44 & 100 & \textbf{5.84} & \textbf{52.05} & 100 & 26.34 & 60.87 & 100\\
& MRHC \textcolor{MRHC_color}{$\blacktriangledown$} && \textbf{40.67} & \textbf{69.34} & 60.80 & 4.67 & 51.43 & 55.55 & \textbf{26.70} & \textbf{61.21} & 59.62\\
\cdashline{2-12}
& MChen$_{10}$ \symMChen && 11.71 & 55.30 & \phantom{0}9.95 & 1.00 & 50.07 & 10.00 & 12.78 & 55.07 & \phantom{0}9.98\\
& MChen$_{30}$ \symMChen && 13.18 & 55.91 & 29.85 & 1.42 & 50.19 & 30.00 & 16.29 & 56.45 & 29.94\\
\cmidrule(lr){1-12}
\multicolumn{2}{l}{\textbf{ML-\textit{k}NN}}\\
& ALL \small{$\blacksquare$} && 29.04 & 64.04 & 100 & \textbf{4.61} & 51.39 & 100 & \textbf{23.71} & 59.99 & 100\\
& MRHC \textcolor{MRHC_color}{$\blacktriangledown$} && 24.58 & 61.77 & 60.80 & 4.24 & 51.16 & 55.55 & 19.75 & 58.17 & 59.62\\
\cdashline{2-12}
& MChen$_{10}$ \symMChen && 15.33 & 56.92 & \phantom{0}9.95 & 1.44 & 50.25 & 10.00 & 13.94 & 55.62 & \phantom{0}9.98\\
& MChen$_{70}$ \symMChen && 22.59 & 60.82 & 69.98 & 3.61 & 51.13 & 70.00 & 20.92 & 58.90 & 69.97\\
& MRSP1$_{30}$ \symMRSPOne && 27.26 & 63.52 & 75.44 & 4.40 & \textbf{51.53} & 80.86 & 23.23 & \textbf{60.15} & 74.51\\
& MRSP2$_{50}$ \symMRSPTwo && \textbf{30.13} & \textbf{64.76} & 83.72 & 3.84 & 51.14 & 85.40 & 23.52 & 60.08 & 83.59\\
\bottomrule[1pt]
\end{tabular}
}
\end{table}

In light of the results obtained, a first remark that may be observed is that the label imbalance in the data severely affects the overall performance of the schemes. More precisely, while the F$^{S}_{1}$ score in the \textit{Moderate} scenario gets to achieve values of up to $40\%$ (MRHC with LP-$k$NN), the same metric rarely surpasses a $5\%$ in the \textit{High} scenario (ALL case with LP-$k$NN). Similarly, the AUC$^{S}$ metric evaluation barely surpasses the random guess (\textit{i.e.}, AUC$^{S} = 50\%$) when addressing highly-imbalanced data whereas the moderately-imbalance corpora generally achieve AUC$^{S}$ values over $60\%$. The \textit{All corpora} case shows an alike behaviour to that of the \textit{Moderate} scenario as the almost-balanced corpora that this assortment incorporates remarkably reduce the overall imbalance ratio. The subsequent analyses thoroughly develop the presented general observations for each of the imbalance-level scenarios.

Focusing on the \textit{Moderate} assortment, it may be checked that the different approaches follow similar trends to those observed in the previous sections: MChen generally depicts the least competitive classification rates due to its sharp reduction whereas the rest of the methods improve these figures as they perform more conservative reduction processes. Moreover, while the ALL case typically represents the best-performing option in terms of classification rate, some of the PG methods do prove to outperform its results---\textit{e.g.}, the MRHC with the LP-$k$NN classifier or the MRSP2$_{30}$ with the ML-$k$NN one. Such an effect is mostly due to the inherent noise cleansing capabilities of the different PG alternatives, which prove to work in these relatively imbalanced scenarios.

In addition, and as previously discussed, these experiments also state a dependency between the reduction technique and the classification strategy. More precisely, MRHC does report the best performance when paired with the LP-$k$NN method whereas the MRSP family is particularly relevant in the BR$k$NN and ML-$k$NN scenarios. Such an insight should be further analysed in future work with the aim of devising a PG strategy that adequately exploits the individual advantages of each $k$NN-based multilabel classification algorithm.

Regarding the \textit{High} imbalance scenario, it can be observed that the classification performance generally degrades after the reduction process. While some of the techniques still report competitive figures---\textit{e.g.}, the case of the ML-$k$NN classifier where the MRSP1$_{30}$ reports a decrease of just $0.21\%$ in the F$^{S}_{1}$ metric with respect to the ALL case or the BR$k$NN scenario in which the MRHC decreases just $0.48\%$ in the F$^{S}_{1}$ score compared to the exhaustive search---it is noticeable that none of the reduced cases outperform the ALL case. The sole exception to this is the case of the MRSP1$_{30}$ strategy that slightly improves the exhaustive search with the ML-$k$NN classifier in the AUC$^{S}$ figure of merit.

Such a performance decrease in these particular imbalance scenarios relates to the fact that no considerations for such cases were contemplated when devising the methods. In this regard, a severely under-represented label may be assumed as noise, being most likely removed from the resulting set. This can be observed in the different MChen cases, in which the performance remarkably degrades, especially when set to perform sharp reductions (\textit{i.e}, low $m$ parameter). Such a limitation is expected to be thoroughly studied with the aim of devising specific multilabel PG policies capable of dealing with class imbalance.

The case of the \textit{All corpora} assortment shows an alike behaviour to that of the \textit{Moderate} imbalance in that the reduction methods generally show slightly lower classification rates than the exhaustive search. An exception to this is the MRSP2$_{70}$ strategy with the BR$k$NN classifier or the MRSP1$_{30}$ and MRSP2$_{50}$ alternatives with the ML-$k$NN model that surpass their respective ALL baseline cases in, at least, one of figures of merit, reinforcing the noise reduction capabilities of the methods. On a final note, it may be observed that, when set to high $m$ values, the MRSP family generally achieves similar classification scores to the exhaustive search, while still performing certain size reduction. In this regard, in the event of addressing a given data collection with an unknown imbalance degree, it may be advisable to consider these PG alternatives in contrast to other possibilities such as the MChen or the reference MRHC methods.

%-------------------------------------------------------------------------------------------------
\subsection{Execution time benchmark}

\begin{figure}[!h]
    \centering
	\includegraphics[width=.9\textwidth]{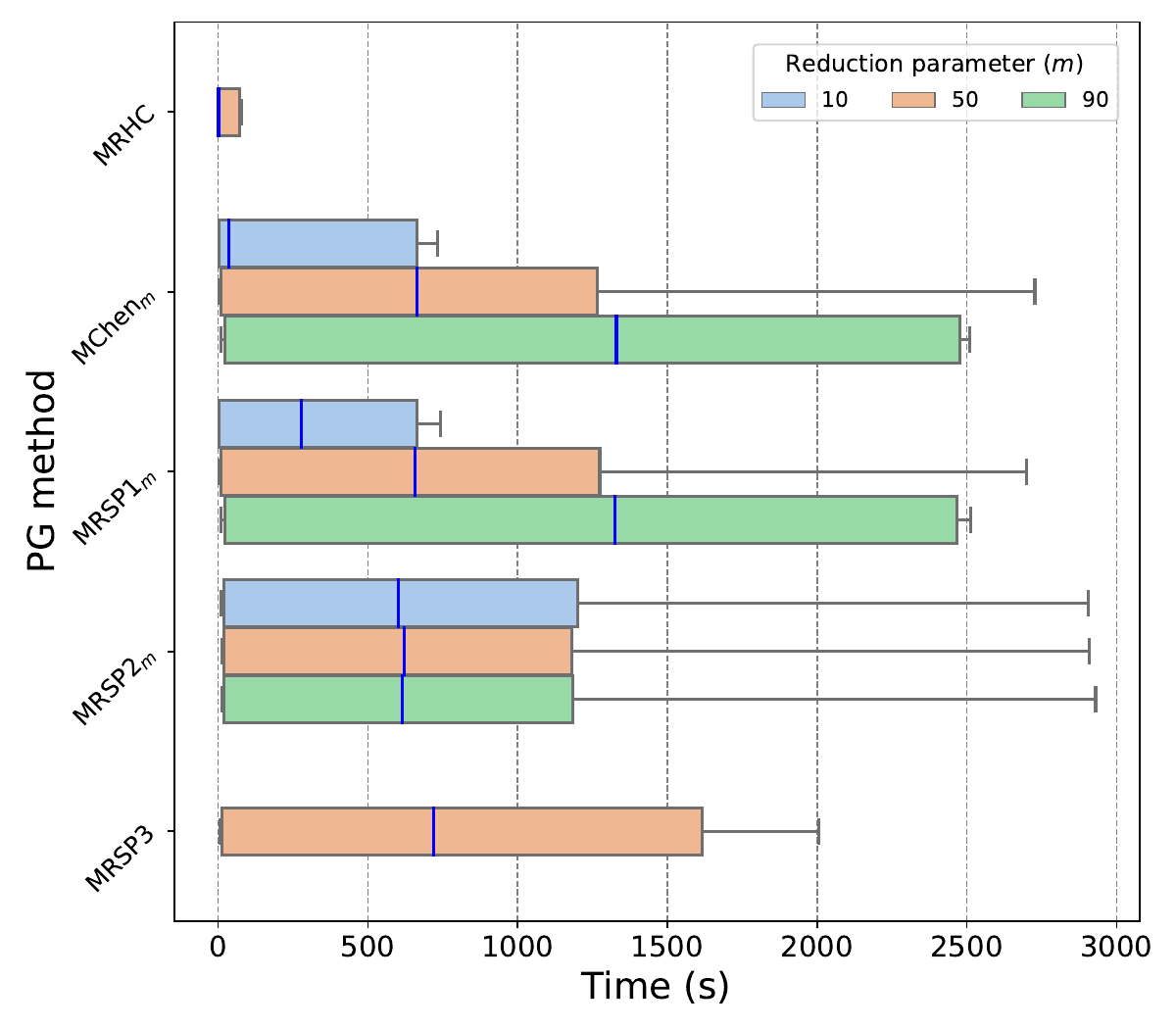}
    \caption{Results, in seconds, of the execution time for each multilabel PG method and $m$ reduction parameter (when applicable). Each sample of the boxplot graph stands for the reduction time obtained in each particular corpus and fold. Note that the cases of $m=30$ and $m=70$ have been omitted for conciseness as they represent intermediate cases of the reported figures.}
    \label{fig:time_benchmarking}
\end{figure}

This last experiment assesses the proposed multilabel PG methods in terms of their execution time and compares them with that of the reference MRHC method. For that, we have performed five different executions of all the reduction processes for each particular algorithm configuration and corpus when addressing the case in which no label noise is induced in the data. The results obtained are summarised in Figure~\ref{fig:time_benchmarking}, where the samples of the boxplot graphs correspond to the individual execution times obtained in each of the aforementioned reduction scenarios. Note that, this evaluation does not relate to the computational complexity of the studied methods but to the efficiency figures of the precise implementations facilitated in the code repository.

As it can be checked, based on the time execution benchmark, the MRHC implementation stands as the most competitive proposal of all the reduction strategies studied. Such a result is totally reasonable since the space partitioning stage of the MRHC algorithm, which is based on the $k$-means clustering method~\cite{hart2000pattern}, has been directly drawn from the optimised and efficient-oriented scikit-learn library~\cite{scikit-learn}. The rest of the implementations, however, do not consider any optimised toolkit that may boost the different procedures within, hence depicting higher execution times.

In a more detailed analysis, it is observed that the execution time for all configurable methods generally grows as the $m$ reduction parameter increases. The sole exception to this assertion is the case of the MRSP2 method, in which the figures obtained remain relatively stable for all $m$ configurations. Taking into account the low computational burden of the MRSP2 prototype merging policy, this stability insight suggests certain independence between the time consumption of the particular space partitioning policy devised for this MRSP2 method and the $m$ reduction parameter.

Moreover, the presence of outliers in some of the methods---more precisely, MChen$_{20}$, MRSP1$_{20}$, and all MRSP2 versions---points out some difficulties when addressing particular cases. Most likely, this is due to a rather time-consuming space partitioning phase---typically, a large amount of pair-wise dissimilarity computations involving high-dimensional instances---as opposed to the prototype merging one, due again to the fact that the merging policies introduced in the work do not entail such elevated computation burdens.

In light of the results obtained, certain mechanisms may be introduced to palliate the inefficiency figures observed, such as the pre-calculation of all dissimilarities or the use of optimised libraries (\textit{e.g.}, scikit-learn) to boost some of the intermediate processes. Nevertheless, and as aforementioned, the observed inefficiency issues as well as the conclusions drawn out of them only relate to the particular implementations facilitated as part of this work.

Finally, it must be highlighted that the figures provided in this analysis exclusively reflect the time invested in reducing the set of training data, which may be deemed as the equivalent to the train phase in an eager learning model. Hence, once this set of data is reduced, the inference time only relates to the inherent efficiency of the $k$NN-based classifier as well as the size of the resulting set obtained by the PG method, but not to the execution time of the latter.

%-------------------------------------------------------------------------------------------------
\section{Conclusions and Future Work}
\label{sec:conclusions}

Prototype Generation (PG) represents one of the most competitive approaches for improving the efficiency of the $k$-Nearest Neighbour ($k$NN) classifier, which is typically related to low-efficiency figures when tackling scenarios with large amounts of data. Nevertheless, while PG methods are commonly considered in multiclass scenarios, very scarce works have addressed such a task in multilabel frameworks.

This work presents the first-time adaptation of four multiclass PG methods to the multilabel case: the reference Chen method~\citep{chen1996sample} and the three versions of the well-known Reduction through Space Partitioning~\citep{sanchez2004high}. For that, we generalise to the multilabel space the different criteria considered by each method for gathering sets of prototypes (space partitioning stage) which are then combined according to certain policies (prototype merging). These novel proposals have been evaluated with 3 multilabel $k$NN-based classifiers, 12 multilabel corpora comprising a varied range of domains and corpus sizes, and different noise scenarios obtained by exchanging the labels of the instances in the train partition.

The results obtained show that the proposed adaptations are capable of significantly improving, both in terms of efficiency and efficacy, the only reference work in the literature---Multilabel Reduction through Homogeneous Clustering method by Ougiaroglou et al.~\cite{Ougiaroglou:MRHC}---as well as the case in which no PG method is applied---the exhaustive search. It is also proved that some of these adaptations show high robustness and data cleansing capabilities in the presence of noise. More precisely, when set to high reduction rates, the proposed Multilabel Chen strategy allows training classification schemes that statistically outperform those trained with the existing baseline approaches. Moreover, the user parameter of these methods allows prioritising either the efficiency or performance features of the scheme, depending on the particular application.

Future work considers the further analysis of the proposed methods contemplating the particularities of multilabel scenarios such as their performance in relation to the data label cardinality or the possible correlations among labels. The second point of interest is the exploration of alternative criteria for the partitioning and prototype merging stages, including the proposal of novel homogeneity policies, which may result in more efficient and/or robust classifiers as well as tackling the limitations observed in these methods when dealing with imbalanced data. Moreover, we consider that additional insights may be obtained by performing other noise induction policies such as swapping only part of the labels between instances, randomly including or eliminating classes for each prototype, or simply duplicating labelsets among elements in the corpus.

In a more practical sense, the presented methods would remarkably benefit from an efficient implementation so that their execution time could be reduced, which represents one of the main limitations observed. From a more general perspective, given the commented scarcity of multilabel PG strategies, future research contemplates the adaptation of other multiclass PG schemes to this particular scenario. Finally, in light of the noise robustness capabilities of the proposals, they may be considered as preprocessing techniques for other classification schemes such as Support Vector Machine or neural models in task-oriented cases such as \textit{music tagging} or \textit{image classification}, among others.

%-------------------------------------------------------------------------------------------------
\section*{Acknowledgments}

This research was partially funded by the Spanish Ministerio de Ciencia e Innovaci{\'o}n through 
the MultiScore (PID2020-118447RA-I00) and 
DOREMI (TED2021-132103A-I00) projects. 
%project MultiScore (PID2020-118447RA-I00) and by the Generalitat Valenciana through project ROMA (GV/2021/064). 
The first author is supported by grant APOSTD/2020/256 from ``Programa I+D+i de la Generalitat Valenciana''.

\bibliographystyle{elsarticle-num} %\biboptions{authoryear}

%\bibliography{main}

\begin{thebibliography}{41}

\expandafter\ifx\csname url\endcsname\relax
  \def\url#1{\texttt{#1}}\fi
\expandafter\ifx\csname urlprefix\endcsname\relax\def\urlprefix{URL }\fi
\expandafter\ifx\csname href\endcsname\relax
  \def\href#1#2{#2} \def\path#1{#1}\fi

\bibitem{hart2000pattern}
P.~E. Hart, D.~G. Stork, R.~O. Duda, Pattern classification, Wiley Hoboken,
  2000.

\bibitem{bishop2006pattern}
C.~M. Bishop, Pattern recognition, Machine Learning 128~(9).

\bibitem{SUYANTO2022116857}
S.~Suyanto, S.~Meliana, T.~Wahyuningrum, S.~Khomsah, A new nearest
  neighbor-based framework for diabetes detection, Expert Systems with
  Applications 199 (2022) 116857.

\bibitem{george2022development}
A.~George, X.~A. Mary, S.~T. George, Development of an intelligent model for
  musical key estimation using machine learning techniques, Multimedia Tools
  and Applications (2022) 1--20.

\bibitem{hancer2021wrapper}
E.~Hancer, I.~Hodashinsky, K.~Sarin, A.~Slezkin, A wrapper metaheuristic
  framework for handwritten signature verification, Soft Computing 25~(13)
  (2021) 8665--8681.

\bibitem{mitchell1997machine}
T.~Mitchell, Machine learning, McGraw Hill Burr Ridge, 1997.

\bibitem{deng2016efficient}
Z.~Deng, X.~Zhu, D.~Cheng, M.~Zong, S.~Zhang, Efficient knn classification
  algorithm for big data, Neurocomputing 195 (2016) 143--148.

\bibitem{GALLEGO2022108356}
A.-J. Gallego, J.~R. Rico-Juan, J.~J. Valero-Mas, Efficient k-nearest neighbor
  search based on clustering and adaptive k values, Pattern Recognition 122
  (2022) 108356.

\bibitem{garcia2015data}
S.~Garc{\'\i}a, J.~Luengo, F.~Herrera, Data preprocessing in data mining,
  Vol.~72, Springer, 2015.

\bibitem{ESCALANTE2016569}
H.~J. Escalante, M.~Graff, A.~Morales-Reyes, Pggp: Prototype generation via
  genetic programming, Applied Soft Computing 40 (2016) 569--580.

\bibitem{triguero2012taxonomy}
I.~Triguero, J.~Derrac, S.~Garcia, F.~Herrera, A taxonomy and experimental
  study on prototype generation for nearest neighbor classification, IEEE
  Transactions on Systems, Man, and Cybernetics, Part C (Applications and
  Reviews) 42~(1) (2012) 86--100.

\bibitem{nanni2011prototype}
L.~Nanni, A.~Lumini, {Prototype reduction techniques: A comparison among
  different approaches}, Expert Systems with Applications 38~(9) (2011)
  11820--11828.

\bibitem{zhang2013review}
M.-L. Zhang, Z.-H. Zhou, A review on multi-label learning algorithms, IEEE
  Transactions on Knowledge and Data Engineering 26~(8) (2013) 1819--1837.

\bibitem{Ougiaroglou:MRHC}
S.~Ougiaroglou, P.~Filippakis, G.~Evangelidis, {Prototype Generation for
  Multi-label Nearest Neighbours Classification}, in: Hybrid Artificial
  Intelligent Systems, Springer International Publishing, Cham, 2021, pp.
  172--183.

\bibitem{ougiaroglou2012efficient}
S.~Ougiaroglou, G.~Evangelidis, Efficient dataset size reduction by finding
  homogeneous clusters, in: Proceedings of the Fifth Balkan Conference in
  Informatics, 2012, pp. 168--173.

\bibitem{GALLEGO2018531}
A.-J. Gallego, J.~Calvo-Zaragoza, J.~J. Valero-Mas, J.~R. Rico-Juan,
  Clustering-based k-nearest neighbor classification for large-scale data with
  neural codes representation, Pattern Recognition 74 (2018) 531--543.

\bibitem{BelloNapolesVanhoofBello:IDA}
M.~Bello, G.~N{\'a}poles, K.~Vanhoof, R.~Bello, On the generation of
  multi-label prototypes, Intelligent Data Analysis 24~(S1) (2020) 167--183.

\bibitem{moyano2018review}
J.~M. Moyano, E.~L. Gibaja, K.~J. Cios, S.~Ventura, Review of ensembles of
  multi-label classifiers: models, experimental study and prospects,
  Information Fusion 44 (2018) 33--45.

\bibitem{GibajaVentura:ACM:Multilabel}
E.~Gibaja, S.~Ventura, {Multi-Label Learning: A Review of the State of the Art
  and Ongoing Research}, Wiley Interdisciplinary Reviews: Data Mining and
  Knowledge Discovery 4~(6) (2014) 411–444.

\bibitem{zhang2018binary}
M.-L. Zhang, Y.-K. Li, X.-Y. Liu, X.~Geng, Binary relevance for multi-label
  learning: an overview, Frontiers of Computer Science 12~(2) (2018) 191--202.

\bibitem{RASTIN2021107526}
N.~Rastin, M.~Z. Jahromi, M.~Taheri, A generalized weighted distance k-nearest
  neighbor for multi-label problems, Pattern Recognition 114 (2021) 107526.

\bibitem{tsoumakas2010random}
G.~Tsoumakas, I.~Katakis, I.~Vlahavas, Random k-labelsets for multilabel
  classification, IEEE Transactions on Knowledge and Data Engineering 23~(7)
  (2010) 1079--1089.

\bibitem{zhang2007ml}
M.-L. Zhang, Z.-H. Zhou, {ML-KNN: A lazy learning approach to multi-label
  learning}, Pattern Recognition 40~(7) (2007) 2038--2048.

\bibitem{younes2008multi}
Z.~Younes, F.~Abdallah, T.~Den{\oe}ux, Multi-label classification algorithm
  derived from k-nearest neighbor rule with label dependencies, in: 2008 16th
  European Signal Processing Conference, IEEE, 2008, pp. 1--5.

\bibitem{cheng2009combining}
W.~Cheng, E.~H{\"u}llermeier, Combining instance-based learning and logistic
  regression for multilabel classification, Machine Learning 76~(2) (2009)
  211--225.

\bibitem{ZHU2021106933}
X.~Zhu, C.~Ying, J.~Wang, J.~Li, X.~Lai, G.~Wang, Ensemble of ml-knn for
  classification algorithm recommendation, Knowledge-Based Systems 221 (2021)
  106933.

\bibitem{chen1996sample}
C.~H. Chen, A.~J{\'o}{\'z}wik, A sample set condensation algorithm for the
  class sensitive artificial neural network, Pattern Recognition Letters 17~(8)
  (1996) 819--823.

\bibitem{sanchez2004high}
J.~S. S{\'a}nchez, High training set size reduction by space partitioning and
  prototype abstraction, Pattern Recognition 37~(7) (2004) 1561--1564.

\bibitem{CastellanosValeroMasCalvoZaragoza:SOCO:PGString}
F.~J. Castellanos, J.~J. Valero-Mas, J.~Calvo-Zaragoza, Prototype generation in
  the string space via approximate median for data reduction in nearest
  neighbor classification, Soft Computing 25 (2021) 15403--15415.

\bibitem{mulan}
G.~Tsoumakas, E.~Spyromitros-Xioufis, J.~Vilcek, I.~Vlahavas, Mulan: A java
  library for multi-label learning, Journal of Machine Learning Research 12
  (2011) 2411--2414.

\bibitem{ScikitML}
P.~{Szyma{\'n}ski}, T.~Kajdanowicz, {Scikit-Multilearn: A Scikit-Based Python
  Environment for Performing Multi-Label Classification}, Journal of Machine
  Learning Research 20~(1) (2019) 209–230.

\bibitem{natarajan2013learning}
N.~Natarajan, I.~S. Dhillon, P.~K. Ravikumar, A.~Tewari, Learning with noisy
  labels, Advances in Neural Information Processing Systems 26.

\bibitem{madjarov2012extensive}
G.~Madjarov, D.~Kocev, D.~Gjorgjevikj, S.~D{\v{z}}eroski, An extensive
  experimental comparison of methods for multi-label learning, Pattern
  Recognition 45~(9) (2012) 3084--3104.

\bibitem{RICOJUAN2019105803}
J.~R. Rico-Juan, J.~J. Valero-Mas, J.~Calvo-Zaragoza, Extensions to rank-based
  prototype selection in k-nearest neighbour classification, Applied Soft
  Computing 85 (2019) 105803.

\bibitem{CALVOZARAGOZA20151608}
J.~Calvo-Zaragoza, J.~J. Valero-Mas, J.~R. Rico-Juan, {Improving kNN
  multi-label classification in Prototype Selection scenarios using class
  proposals}, Pattern Recognition 48~(5) (2015) 1608--1622.

\bibitem{valero2017experimental}
J.~J. Valero-Mas, J.~Calvo-Zaragoza, J.~R. Rico-Juan, J.~M. I{\~n}esta, An
  experimental study on rank methods for prototype selection, Soft Computing
  21~(19) (2017) 5703--5715.

\bibitem{demvsar2006statistical}
J.~Dem{\v{s}}ar, Statistical comparisons of classifiers over multiple data
  sets, Journal of Machine Learning Research 7 (2006) 1--30.

\bibitem{KanjAbdallahDenoeuxTout:TPAMI:2016}
S.~Kanj, F.~Abdallah, T.~Denoeux, K.~Tout, Editing training data for
  multi-label classification with the k-nearest neighbor rule, Pattern Analysis
  and Applications 19~(1) (2016) 145--161.

\bibitem{ArnaizGonzalezDiezPastorRodriguezGarciaOsorio:ASOC:2018}
Álvar Arnaiz-González, J.-F. Díez-Pastor, J.~J. Rodríguez,
  C.~García-Osorio, Local sets for multi-label instance selection, Applied
  Soft Computing 68.

\bibitem{LiuBlekasTsoumakas:PR:2022}
B.~Liu, K.~Blekas, G.~Tsoumakas, Multi-label sampling based on local label
  imbalance, Pattern Recognition 122 (2022) 108294.

\bibitem{scikit-learn}
F.~Pedregosa, G.~Varoquaux, A.~Gramfort, V.~Michel, B.~Thirion, O.~Grisel,
  M.~Blondel, P.~Prettenhofer, R.~Weiss, V.~Dubourg, J.~Vanderplas, A.~Passos,
  D.~Cournapeau, M.~Brucher, M.~Perrot, E.~Duchesnay, {Scikit-learn: Machine
  Learning in Python}, Journal of Machine Learning Research 12 (2011)
  2825--2830.

\end{thebibliography}

% -------------------------------------------------------------------
% Authors' bios

\vspace{0.5cm}

\textbf{Jose J. Valero-Mas} obtained the M.Sc. in Telecommunications Engineering from the University Miguel Hernández of Elche in 2012, the M.Sc. in Sound and Music Computing from the Universitat Pompeu Fabra in 2013, and the Ph.D. in Computer Science from the University of Alicante in 2017. He is currently a postdoctoral researcher with a grant from the Valencian Government at the Department of Software and Computing Systems of the University of Alicante, Spain. His research interests include Pattern Recognition, Machine Learning, Music Information Retrieval, and Signal Processing for which he has co-authored more than 30 works within international journals, conference communications, and book chapters.

\vspace{0.5cm}
\textbf{Antonio Javier Gallego} is an associate professor in the Department of Software and Computing Systems at the University of Alicante, Spain. He received B.Sc. \& M.Sc. degrees in Computer Science from the University of Alicante in 2004, and a Ph.D. in Computer Science and Artificial Intelligence from the same university in 2012. He has been a researcher on 15 research projects funded by the Spanish Government and private companies. He has authored more than 60 works published in international journals, conferences, and books. His research interests include Deep Learning, Pattern Recognition, Computer Vision, and Remote Sensing.

\vspace{0.5cm}
\textbf{Pablo Alonso-Jim{\'e}nez} holds a B.Sc. in Telecommunications Engineering from the Universidad de Vigo (2016) and an M.Sc. in Sound and Music Computing from the Universitat Pompeu Fabra (2017), where he is currently pursuing the Ph.D. in the Music Technology Group under the supervision of Dr. Xavier Serra. His fields of interest include signal processing and machine learning for audio, speech, and music applications.

\vspace{0.5cm}
\textbf{Xavier Serra} received a Ph.D. degree in computer music from Stanford University, Stanford, CA, USA, in 1989. He is currently a Professor with the Department of Information and Communication Technologies and the Director of the Music Technology Group, Universitat Pompeu Fabra, Barcelona, Spain. His research interests include the computational analysis, description, and synthesis of sound and music signals. Dr. Serra is very active in the fields of audio signal processing, sound and music computing, music information retrieval and computational musicology at the local and international levels, being involved in the editorial board of a number of journals and conferences and giving lectures on current and future challenges of these fields. He was awarded an Advanced Grant from the European Research Council to carry out the project CompMusic aimed at promoting multicultural approaches in music information research.

\end{document}